\documentclass[preprint,5p]{elsarticle}

\usepackage{amssymb,amsmath,amsthm}
\usepackage{graphicx,epstopdf,color}
\usepackage{algorithm}
\usepackage{algpseudocode}
\usepackage{caption}
\usepackage{subcaption}
\usepackage{wrapfig}

\journal{Neurocomputing}

\begin{document}

\begin{frontmatter}

\title{Enhancements of Multi-class Support Vector Machine Construction from Binary Learners using Generalization Performance}

\author[cu]{Patoomsiri Songsiri}
\ead{patoomsiri.s@student.chula.ac.th}
\author[wu]{Thimaporn Phetkaew}
\ead{pthimapo@wu.ac.th}
\author[cu]{Boonserm Kijsirikul\corref{cor1}}
\ead{boonserm.k@chula.ac.th}

\address[cu]{Department of Computer Engineering, Chulalongkorn University, Pathumwan, Bangkok Thailand. 10330}
\address[wu]{School of Informatics, Walailak University, Thasala District, Nakhon Si Thammarat Thailand. 80161}

\cortext[cor1]{Corresponding author. Tel.:+66-(0)2-218-6956\\fax: +66-(0)2-218-6955.}

\author{}

\address{}

\begin{abstract}

We propose several novel methods for enhancing the multi-class SVMs by applying the generalization performance of binary classifiers as the core idea. This concept will be applied on the existing algorithms, i.e., the Decision Directed Acyclic Graph (DDAG), the Adaptive Directed Acyclic Graphs (ADAG), and Max Wins. Although in the previous approaches there have been many attempts to use some information such as the margin size and the number of support vectors as performance estimators for binary SVMs, they may not accurately reflect the actual performance of the binary SVMs. We show that the generalization ability evaluated via a cross-validation mechanism is more suitable to directly extract the actual performance of binary SVMs. Our methods are built around this performance measure, and each of them is crafted to overcome the weakness of the previous algorithm. The proposed methods include the Reordering Adaptive Directed Acyclic Graph (RADAG), Strong Elimination of the classifiers (SE), Weak Elimination of the classifiers (WE), and Voting based Candidate Filtering (VCF). Experimental results demonstrate that our methods give significantly higher accuracy than all of the traditional ones. Especially, WE provides significantly superior results compared to Max Wins which is recognized as the state of the art algorithm in terms of both accuracy and classification speed with two times faster in average. 

\end{abstract}


\end{frontmatter}

\section{Introduction}\label{introduction}

The support vector machine (SVM)~\cite{Vapnik98,Vapnik99} is a high performance learning algorithm
constructing a hyperplane to separate two-class data by maximizing the margin between them.
There are two approaches for extending SVMs to multi-class problems, i.e.,
solving the problem by formulating all classes of data under a single optimization,
and combining several two-class subproblems. However, the difficulty and complexity
to solve the problem with the first method are due to the increase of the number
of classes and the size of training data, so the second method is more suitable for practical use.
In this paper, we focus on the second approach.  

For constructing a multi-class classifier from binary ones, the method called {\it one-against-one} trains each binary classifier on only two out of $N$ classes, and builds $N(N-1)/2$ possible classifiers. Several strategies have been proposed for combining the trained classifiers to make the final classification for an unseen data. Friedman~\cite{Friedman96} suggested the combination strategy called {\it Max Wins}. In the classification process of Max Wins, every binary classifier provides one vote for its preferred class and 
the class with the largest vote will be set to be the final output.
Chang and Lee ~\cite{Chang11} investigated an adaptive framework to manage 
a {\it nuisance} vote which is a vote for an unrelated class by allowing a classifier to make a non-vote for data of unrelated class. Instead of a binary classifier, they employed a ternary 
classifier that consists of two particular classes and 
the rest of the classes fused as the third class for solving this problem.

Vapnik~\cite{Vapnik98} proposed the one-against-the-rest approach working by constructing a set of $N$ binary classifiers
in which each $i^{th}$ classifier is learned from all examples in the $i^{th}$ class, and the remaining classes labeled  
with the positive and negative classes, respectively.
The class corresponding to the classifier with the highest output value
is used to make the final output.
Moreover, Manikandan and Venkataramani~\cite{Manikandan10} adapted the traditional one-against-the-rest
to work as a sequential classifier. All classifiers will be ordered corresponding 
to their misclassification.
This method needs a lower number of classifiers on avearge 
compared with the traditional one-against-the-rest, 
but both algorithms have the same problem in the training phase 
because of the difficulty for calculating the absolutely separating hyperplane 
between a class and all of the other classes.

Dietterich and Bakiri~\cite{Dietterich95} introduced the Error Correcting Output Code (ECOC)
based on the fundamental of information theory.
For a given code matrix with $N$ rows and $L$ columns, each element contains either 
$\textquoteleft$1', or $\textquoteleft$-1'.
Each column denotes the bit string showing the combination 
of positive and negative classes for constructing a binary classifier, 
and each row of the code matrix indicates the unique bit string for representing a specific class 
(each bit string is called a {\it{codeword}}). 
Allwein et al.~\cite{Allwein2000} extended the coding method by adding the third symbol $\textquoteleft$0'
as {\it`` don't care bit''} to allow the binary model learned without considering some particular classes.
Unlike the previous method, the number of classes for training a binary classifier 
can be varied from 2 to $N$ classes. 
Based on these two systems for an $N$ classes problem, the maximum numbers of 
different binary classifiers are $2^{N-1}-1$~\cite{Dietterich95}, 
and $\frac{3^{N}-2^{N+1}+1}{2}$~\cite{Bagheri2013}, respectively.
Design of code matrices with different subsets of binary classifiers
gives different abilities for separating classes, and the problem of selecting a suitable subset of binary classifiers 
is complicated with a large size of $N$. 
To obtain the suitable code matrix,  some techniques using the Genetic Algorithm have been proposed~\cite{Kuncheva2005,Lorena2009}.
In the classification phase, a test example is classified by all classifiers corresponding to the column of the code matrix,
and then the class with the closet codeword is assigned to the final output class.

Platt et al.~\cite{Platt99} proposed the Decision Directed Acyclic Graph (DDAG)
in order to reduce evaluation time ~\cite{Hsu02}. In each round,
a binary model will be randomly selected from all $N(N-1)/2$ classifiers.
The binary classification result is employed to eliminate the candidate output classes, and
to ignore all binary classifiers related to the defeated class. 
It guarantees that the number of classifications (applied classifiers) of the DDAG is always $N-1$.
This recursive task will be applied until there is only one last candidate class.
However, the misclassification of the DDAG can be occurred at the time when 
selected {\it binary classifiers related to the target class} (hence forth {\it BCRT})
give the wrong answer.
The more times the number of BCRTs are applied,
the more chance the misclassification is produced by the DDAG.  
In order to reduce this risk, Kijsirikul and Ussivakul~\cite{Kijsirihul02} proposed the
Adaptive Directed Acyclic Graphs (ADAG) that has a reversed triangular structure of the DDAG. 
It requires only $\lceil log_2N \rceil$ times or less that the target class is tested against the other classes,
while the DDAG possibly requires at most $N-1$ times.

In addition, there have been many attempts that apply some information such as 
the margin size~\cite{Platt99}, the number of support vectors~\cite{Abe03}, 
and the separability measures between classes~\cite{Lorena11,Li11}, 
to improve the performance of the multi-class classification.
The margin size and the number fo support vectors were applied for selecting the suitable two-class classifiers in the DDAG~\cite{Platt99,Abe03}.
The separability measure was employed for automatically constructing 
a binary tree of multi-class classification based on the concept of the minimum spanning tree~\cite{Lorena11}.
Li, et al.~\cite{Li11} used similar information to vote the preferred class for data in unclassifiable region 
for both the one-against-one and the one-against-the-rest techniques.

In this research, we investigate the framework for enhancing three well-known methods, 
which are the DDAG, the ADAG, and Max Wins.
Max Wins is currently recognized as the-state-of-the-art combining algorithm and 
it is also the most powerful technique among all of our focused works with 
a need of $N(N-1)/2$ number of classifications for an $N$-class problem,
while the other two approaches reduce the number of classifications to $N-1$.
We study the characteristics of these methods
that lead to wrong classification results. The first two techniques have the same hierarchical structure and have 
the same weak point that they {\it``trust on individual opinion''} 
for making decision to discard the candidate classes. 
Intuitively, if only one of BCRTs makes a mistake, 
the whole system will give the wrong output. 
The last technique as the high performance one, Max Wins is based on the concept of {\it``trust on most popular opinion''} 
for making decision to select the output class. 
If all of $N-1$ BCRTs give the correct answer, Max Wins will always provide the correct output class. 
However, if there exists only one of BCRTs give the wrong answer,  
it may lead to misclassification due to equal voting, or other non-target classes 
reaching the largest vote as shown later in the paper.
Examples which are incorrectly classified in this scenario can be recovered by our proposed strategies.

In this paper, we demonstrate that the above traditional methods can be improved based on the same idea that if we access further important information of {\it generalization performance} of all binary classifiers and properly estimate it, it can be employed for enhancing the performance of the methods. 
Based on this idea, we propose four novel approaches including (1) the Reordering Adaptive Directed Acyclic Graph (RADAG),
(2) Strong Elimination of the classifiers (SE), (3) Weak Elimination of the classifiers (WE),
and (4) Voting based Candidate Filtering (VCF). 
The first approach, the next two approaches, and the last approach are improved from
the ADAG, the DDAG, and Max Wins, respectively. 
We also empirically evaluate our methods by comparing them with the traditional methods 
on the sixteen datasets from the UCI Machine Learning Repository~\cite{Blake98}.

This paper is organized as follows. Section~\ref{MSVM} reviews the traditional multi-class classification frameworks.
Section~\ref{Estimate_Generr} describes how to properly estimate the generalization performance of binary classifiers.
Section~\ref{Proposed_Method} presents our proposed methodologies.
Section~\ref{experiments} performs experiments and explains the results and discussions.
Section~\ref{conclusion} concludes the research.

\section{Multi-class Support Vector Machines} \label{MSVM}

\subsection{Max Wins} 
\begin{figure}[htbp]

\centering
\scalebox{1.0}{\includegraphics{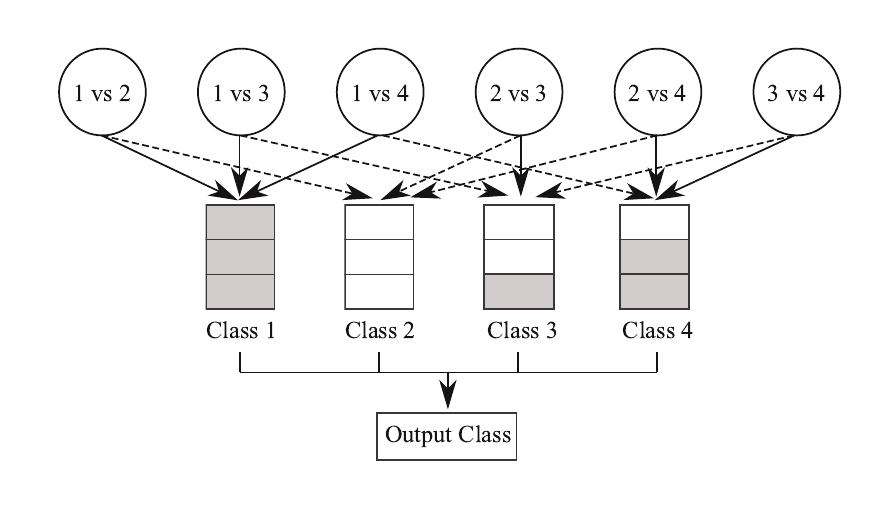}}
\vspace{-30pt}
\caption{An example of a four-class classification with Max Wins.}
\label{maxwin_chart}
\end{figure}

For an $N$-class problem, all possible pairs of two-class data are learned for constructing
$N(N-1)/2$ classifiers. All binary classifiers are applied for voting the preferred class. 
A class with maximum vote will be assigned as the final output class.
This method is called Max Wins~\cite{Friedman96}.
However, in case that there exists more than one class giving the same maximum vote, 
the final output class can be obtained by random selection from candidate classes with the equal maximum-vote.
An example of the classification using this technique for a four-class problem is shown in Fig.~\ref{maxwin_chart}. 
Each class will be voted (solid-line) or ignored (dash-line) by all related binary models.
For example, class $1$ has three related classifiers, i.e., 1 vs 2, 1 vs 3, and 1 vs 4.
The voting result of class $1$, class $2$, class $3$ and class $4$ are three, zero, one, and, two, respectively. 
In this case, class $1$ has the largest score, and therefore it is assigned as the final output class.

\subsection{Decision Directed Acyclic Graphs}

Platt et al. ~\cite{Platt99} introduced a learning algorithm 
using the Directed Acyclic Graph (DAG) to represent the classification task, 
called the Decision Directed Acyclic Graph (DDAG).
This architecture represents a set of nodes connected by edges with no cycles. Each 
edge has an orientation and each node has either 0 or 2 edges.
Among these nodes, there exists a root node which is the unique node with no edge pointing into it. 
In a DDAG, the nodes are arranged with a triangular shape in which each node is labeled with an element of a boolean function. 
There exists a single root node at the top, two nodes in the second layer, and so on until the final layer of $N$ leaves for an $N$-class problem.

\begin{figure}[htbp]
\centering
\scalebox{1.0}{\includegraphics{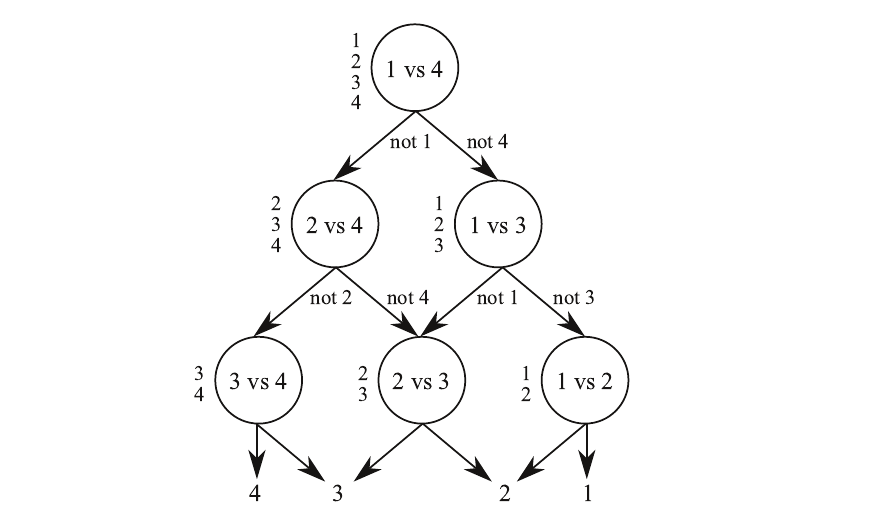}}
\vspace{-20pt}
\caption{The DDAG finding the best class out of four classes ~\cite{Platt99}.}
\label{DDAG}
\end{figure}

To make a classification, an example with an unknown class label is evaluated by the nodes as binary decision functions.
The binary output result in each layer is applied to eliminate the candidate output classes
and the binary classifiers related to the defeated class are removed.
At the first layer (see~Fig.~\ref{DDAG}), the root node can be randomly selected from all possible $N(N-1)/2$ classifiers
and there are $N$ candidate output classes.
After the root node is tested, its binary result is employed to eliminate the candidate output classes
and the binary classifiers corresponding to the defeated class are discarded.
In the next layer, the remaining binary classifiers are randomly selected to continue the same process in which some classes are eliminated from the remaining candidate classes. The process is repeated 
until there is only one class remained which is then assigned as the final output class. 
This algorithm requires only $N-1$ decision nodes in order to obtain the final answer. 

One disadvantage of the DDAG is that its classification result is affected by the sequence of binary classifiers randomly selected 
in the evaluation process. 
Platt et al. also proposed the other method that prefers the binary decision function
with the higher generalization performance measured by its margin sizes, called the large margin DAGs~\cite{Platt99}. 
The margin size ($\Delta$) is a parameter for bounding the generalization ability 
of the binary SVM as shown in terms of the VC-dimension in Eq.~(\ref{SRM_function}).
It illustrates that the generalization performance of the binary model is proportional to
the size of the margin. A binary classifier with the larger margin size will be firstly applied 
in each round of the evaluation step.
Moreover, Takahashi and Abe~\cite{Abe03} proposed a similar framework that
employed the number of support vectors as a performance measure.
In this method, the generalization error ($\epsilon_{ij}$) for classes $i$ and $j$ was bounded by
Eq.~(\ref{gen_err_bounding_by_SV})~\cite{Vapnik95}:

\begin{equation}
\epsilon_{ij}=\frac{SV_{ij}}{M_{ij}},
\label{gen_err_bounding_by_SV}
\end{equation}

\noindent
where $SV_{ij}$ is the number of support vectors for classes $i$
and $j$ and $M_{ij}$ is the number of training data for classes $i$ and $j$.\

\subsection{Adaptive Directed Acyclic Graphs}\label{SectionADAG}

\begin{figure}[htbp]
\centering
\scalebox{0.9}{\includegraphics{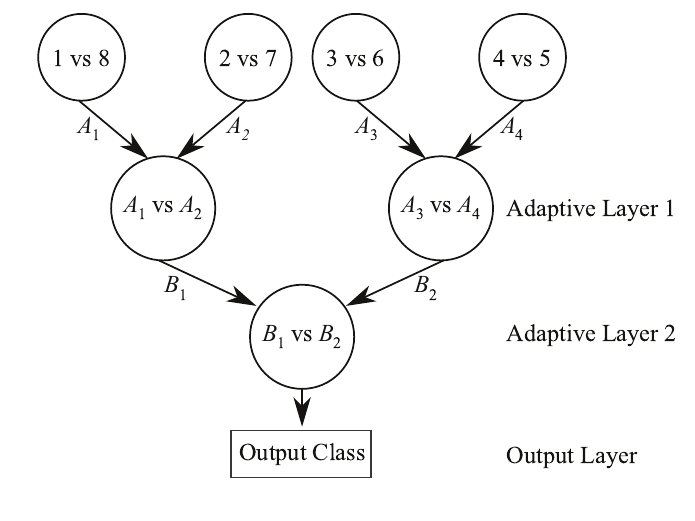}}
\vspace{-10pt}
\caption{The structure of an adaptive DAG for an 8-class problem.}
\label{structureOfADAG}
\end{figure}

In the DDAG, binary classification result of a previously employed binary classifier is used 
to eliminate a candidate output classe, and there are only current remaining candidate classes 
that can be possibly assigned as the final output class. Therefore, the misclassification of a selected 
BCRT is the crucial point. 

The ADAG was originally designed to reduce this risk of the DDAG 
by using reversed triangular structure~\cite{Kijsirihul02}. 
In an $N$-class problem,
there are $\lceil \frac{N}{2} \rceil$ nodes at the top, $N/2^2$ nodes
in the second layer and so on until the lowest layer of the final node,
as illustrated in Fig.~\ref{structureOfADAG}.
Like the DDAG, binary output results of the ADAG in each layer are applied to discard candidate output classes
and the binary classifiers related to the defeated classes are also ignored.
Therefore, the ADAG also evaluates only $N-1$ nodes to obtain the final answer.

According to the critical issue of misclassification mentioned above, 
even only one selected classifier related to the target class
provides a wrong answer, the misclassification on the final output class cannot be avoided. 
Hence, the number of times the target class is tested against other classes
indicates the risk of misclassification.
The DDAG requires at most $N-1$ times that the target class is tested against other classes, 
while the ADAG requires only
$\lceil log_2N \rceil$ times or less. 
This shows that the opportunities of the target class tested against other classes
on the ADAG is much lower than the DDAG. 
 

\section{An Estimation of the Generalization Performan-ces of Binary Support Vector Machines }\label{Estimate_Generr}

The generalization performance of a learning model is the actual performance evaluated on unseen data. 
For support vector machines, a model is trained by using the concept of
the Structure Risk Minimization principle~\cite{Vapnik74} in which the generalization performance 
of the model is estimated based on both terms of the complexity of model (the VC dimension of approximating functions) 
and the quality of fitting training data (empirical error). 
Consider the problem of binary classification where dataset $X$ of $m$ samples
in real $n$-dimensional space is randomly independent identically distributed observations drawn according to
$P(x,y)=P(x)P(y|x)$. The expected risk ($R(\alpha)$) with probability at least $1-\delta$ can be bounded by the following equation~\cite{Bartlett99,Burges98}:

\begin{equation}
R(\alpha) \leq \frac{l}{m}+\sqrt{\frac{c}{m}({\frac{R^2}{\Delta^2} { log^2m } +log{\frac{1}{\delta}}})},
\label{SRM_function}
\end{equation}

\noindent
where there is a corresponding constant $c$ for all probability distributions, 
$l$ is the number of labeled examples in $z$ with margin less than $\Delta$,  
$R$ indicates the radius of the smallest sphere that contains all the data points,
and $\Delta$ is the distance between the hyperplane and the closest points of the training set (margin size).
The first and second terms of inequality in Eq.~(\ref{SRM_function}) denote the bound of the empirical error, 
and the VC dimension, respectively.

In our frameworks, the generalization ability will be applied to improve the multi-class classification.
Although there have been many attempts to use some performance measures
such as the margin size~\cite{Platt99}, the number of support vectors~\cite{Abe03}, 
they may not accurately reflect the actual performance of each binary SVM. 
Consider a two-class problem where hyperplanes $h_1$ and $h_2$ are learning models
created to separate the positive and the negative examples. Suppose that they provide different margin sizes 
of $\Delta_1$ and $\Delta_2$, and the different numbers, {\it $l_1$ and $l_2$,} of labeled examples in $z$ with the margin
less than their margin sizes, respectively. 
In case that the parameters $c$ and $\delta$ are fixed, there are only two parameters 
including $\Delta$ and $l$ that affect the performance of the learning model (as the parameters $m_1$ and $m_2$, as well as $R_1$ and $R_2$ are the same for the same pair of a two-class problem).
Now consider two-learning models learned from different pairs of a two-class problem.
In case that the parameters $c$, and $\delta$ are fixed,
according to inequality in Eq.~(\ref{SRM_function}), obviously, if we use only $\Delta$, $\l$, or combination of them, 
they are not sufficient to represent the whole term of their generalization abilities.
This shows that a binary model with the larger margin size may not provide 
more accurate result of classification. The use of only the number of support vectors is also shown in~\cite{Burges98} that it is not predictive for generalization ability. 

\begin{figure*}
		\vspace{-16pt}
        \centering
        \begin{subfigure}[b]{0.44\textwidth}      
				\centering                          
                \includegraphics[width=\textwidth]{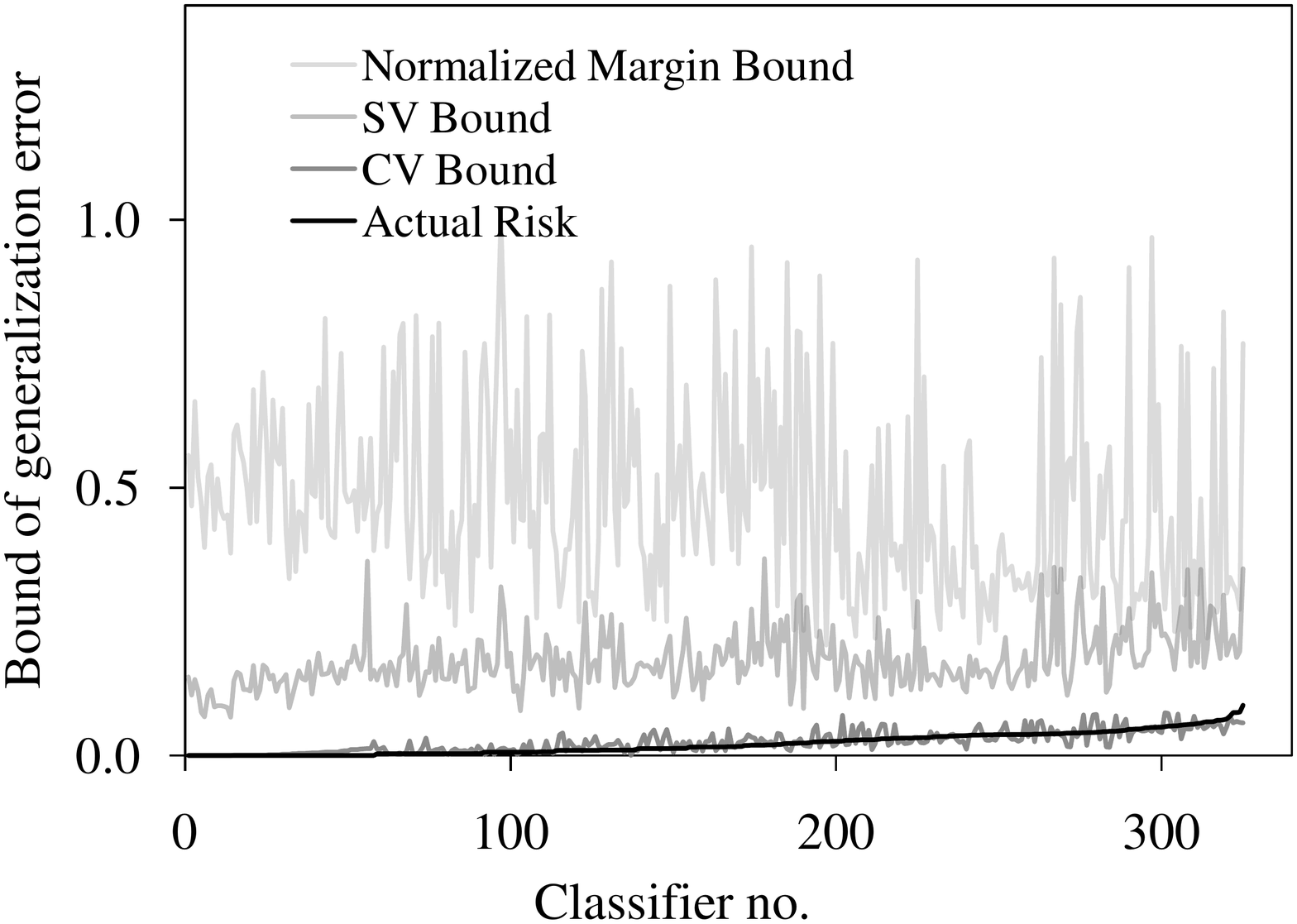}
                \vspace{-26pt}
                \caption{}
                \label{fig:fig04a}
        \end{subfigure}%
        \begin{subfigure}[b]{0.44\textwidth}
				\centering
                \includegraphics[width=\textwidth]{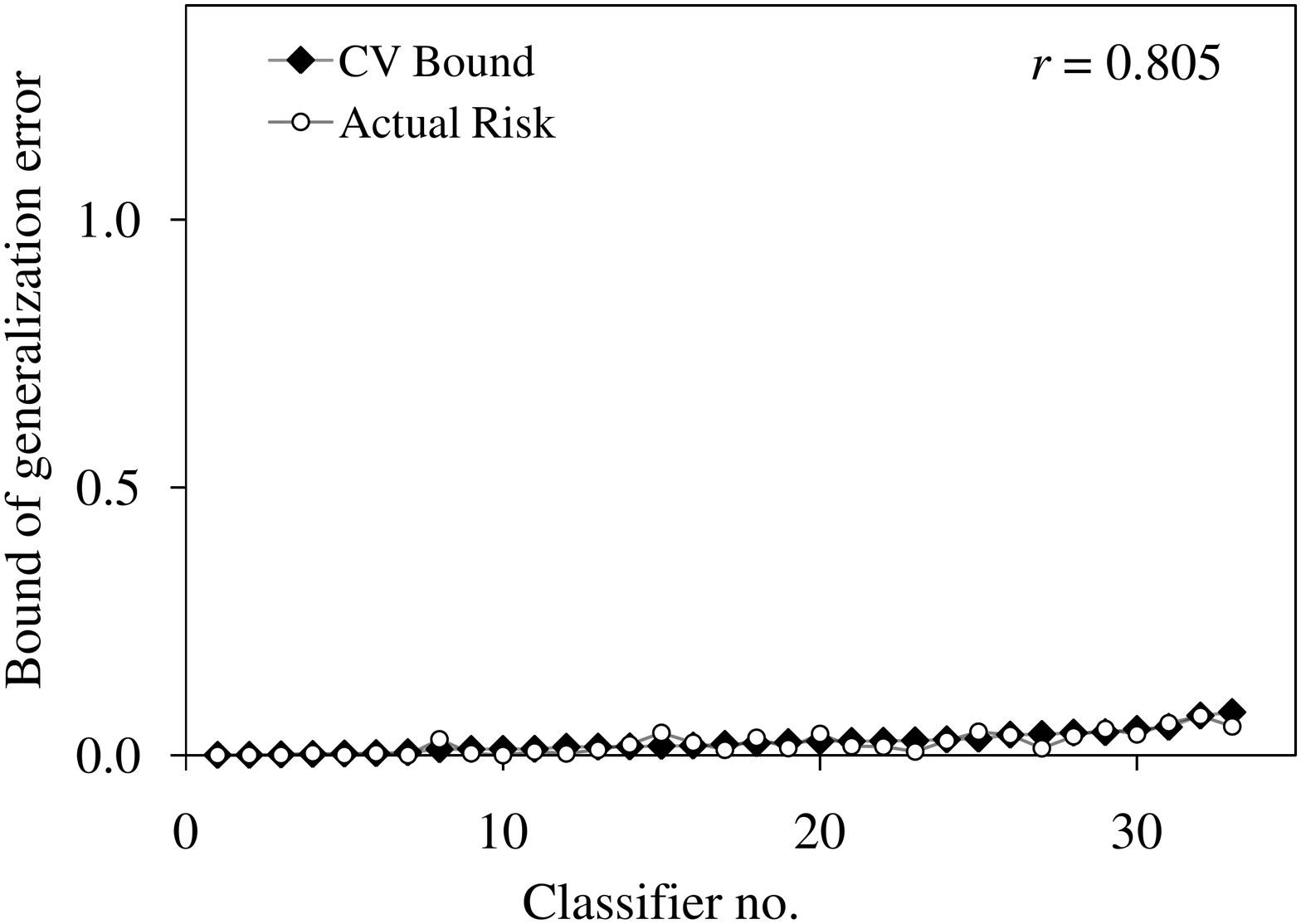}
                \vspace{-26pt}
                \caption{}
                \label{fig:fig04b}
        \end{subfigure}     
        \begin{subfigure}[b]{0.44\textwidth}
				\centering
				\vspace{5pt}
                \includegraphics[width=\textwidth]{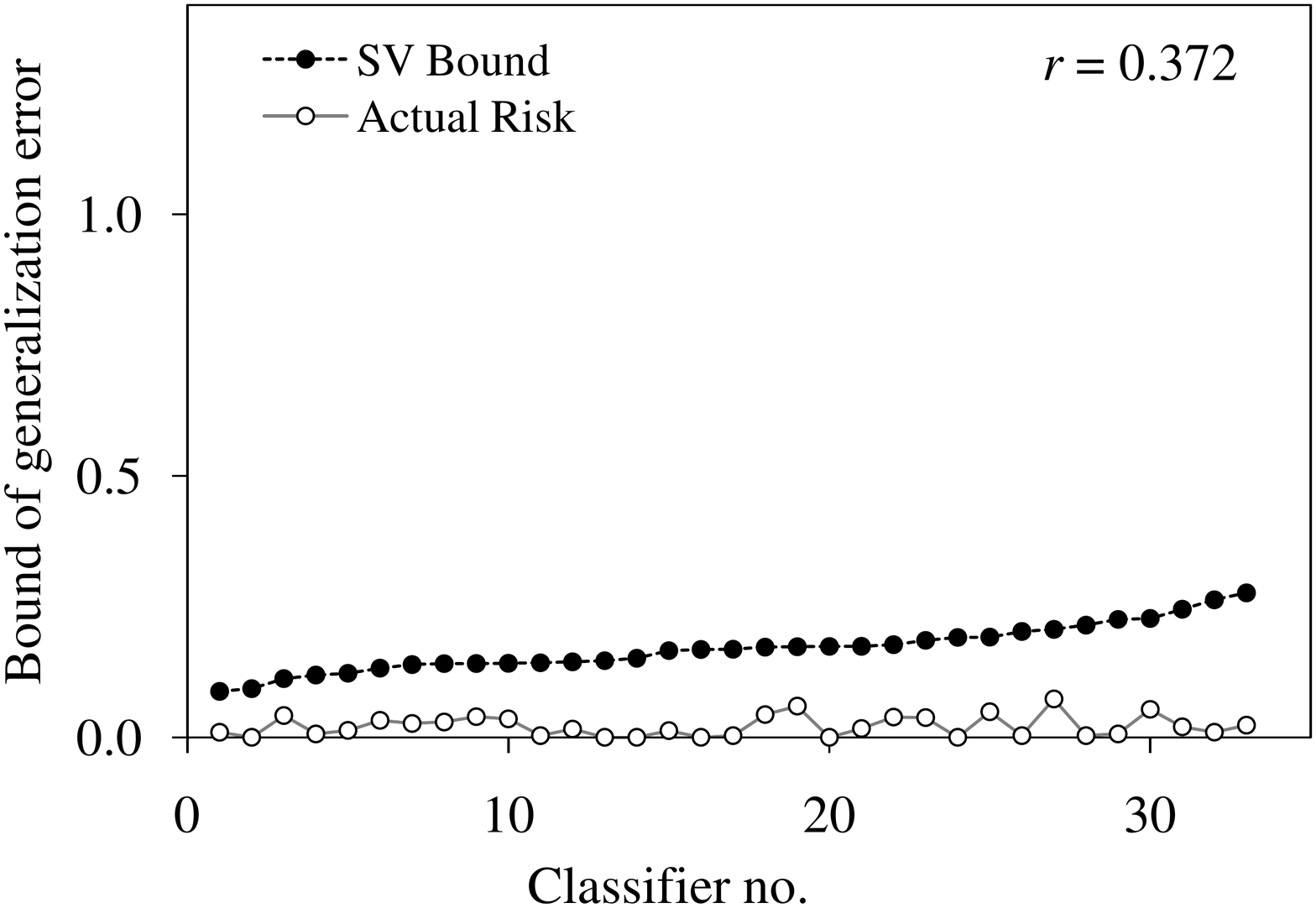}
                 \vspace{-26pt}
                \caption{}
                \label{fig:fig04c}
        \end{subfigure}
        \begin{subfigure}[b]{0.44\textwidth}                
				\centering
				\vspace{5pt}
                \includegraphics[width=\textwidth]{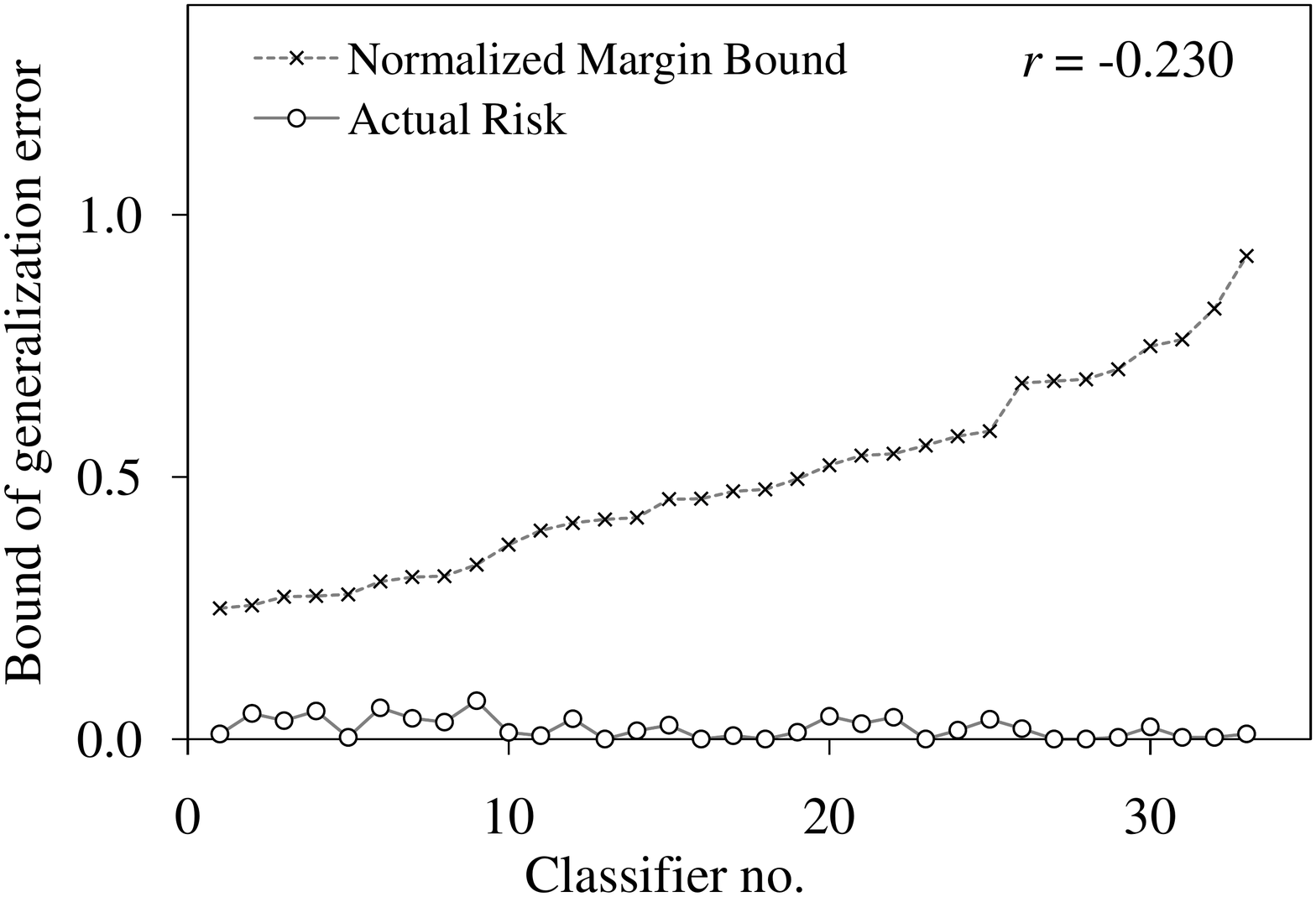}
                 \vspace{-26pt}
                \caption{}
                \label{fig:fig04d}
        \end{subfigure}      
        \vspace{-4pt}		
        \caption{Generalization errors of 325 classifiers of the Letter dataset based on $k$-fold cross-validation (CV Bound), 
the number of support vectors (shown in term of the ratio between the number of support vectors and the number of training data: SV Bound),
the margin size (shown in term of its inverse value normalized to be in [0,1]: Normalized Margin Bound), 
and their actual risks on test data (unseen data) by applying the polynomial kernel of $d=4$. Figure (a) compares generalization errors calculated by all techniques where classifiers are sorted in the ascending order 
by their actual generalization performances (actual risk), and figures (b)-(d) show the comparisons between the actual risks, and the estimated generalization errors with different measures, i.e., CV Bound, SV Bound, and Normalized Margin Bound, respectively (the classifiers will be sorted in ascending order 
by the estimated generalization errors, and for ease of visualization we show only $10\%$ of classifiers 
by sampling every ten classifers  from the sorted list of the classifiers.}\label{gen_err_bound_all}
\end{figure*}

\begin{figure*}[htbp]
\centering
\scalebox{0.95}{\includegraphics{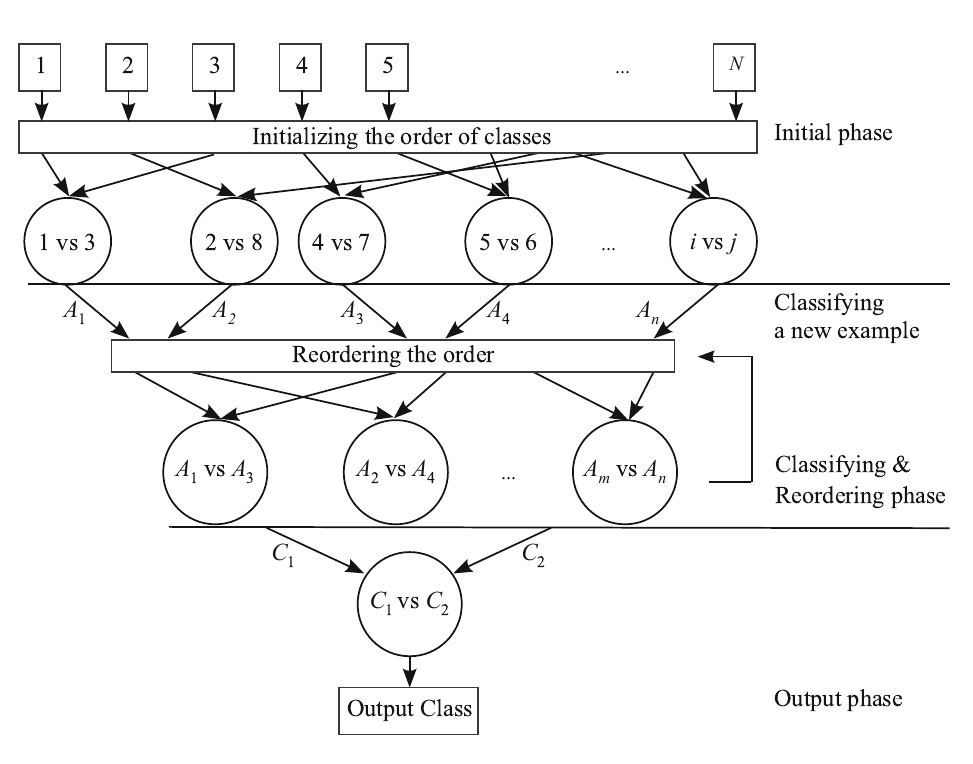}}
\vspace{-10pt}
\caption{Classification process of the RADAG.}
\label{RADAG}
\end{figure*}

As described above, the generalization ability can be employed to enhance the performance of multi-class classification, by carefully design algorithms which utilize this information as a selection measure for good classifiers. We believe that  the generalization performance of binary SVMs can be directly estimated by $k$-fold cross-validation ~\cite{Mitchell97} (see Algorithm~\ref{CV_algo}), and it can be used to fairly compare the performances of binary SVMs on different two-class problems. 
Below we give an example which demonstrates that $k$-fold cross-validation 
is more suitable for estimating the generalizaiton performance of the classifiers 
than the other measures used by the previous methods, i.e. the number of support vectors, the margin size.

\begin{algorithm}[t]
  \caption{An estimation of the generalization error of a classifier by using $k$-fold cross-validation.}\label{CV_algo}
  \begin{algorithmic}[1]
  \Procedure{Cross Validation}{}
  \State Set of training data $T$ is partitioned into $k$ disjoint 
		\hspace*{1.2em} equal-sized subsets
   \State Initial the classification error of round $i$: $\epsilon_i \gets 0$ 
  \For{$i$=1 to $k$ }
     \State {\it validate set} $\gets$ $i^{th}$ subset  
     \State {\it training set} $\gets$ all remaining subsets
     \State Learn model based on {\it training set}
     \State $\epsilon_i$ $\gets$ Evaluate the learned model 
								by {\it validate set}, and find the number of examples with misclassification       
	 	 
  \EndFor  
  \State {\it generalization error} $\gets$ $\sum_{i=1}^{k}\epsilon_i \times \frac{1}{|T|}$ 
  \State \textbf{return} {\it generalization error}
  \EndProcedure
  \end{algorithmic}
\end{algorithm}
 
Fig.~\ref{gen_err_bound_all} shows the generalization performance measured by the previous methods ~\cite{Platt99,Abe03}, and $k$-fold cross-validation, which we propose to use as the performance measure, for the Letter dataset with 26 classes, by applying the polynomial kernel of $d=4$. Fig.~\ref{gen_err_bound_all} (a) illustrates that the trend of estimated generalization error by $k$-fold cross-validation
is very closed to the actual risk, while the other two techniques give high variation. 
To further investigation in more details, we select about $10\%$ of all classifiers to show in Fig.~\ref{gen_err_bound_all} (b-d); these figures
illustrate the comparisons between the actual risk and the estimated generalization errors with different measures, i.e., CV Bound, SV Bound, and Normalized Margin Bound, respectively. In each figure, classifiers are sorted in ascending order by the estimated generalization errors. 
It is expected that if a specific measure is a good estimator for generalization error, its value should be in the same trend as the actual risk (its value shoud increase with the increase of the actual risk). A good trend is found in Fig.~\ref{gen_err_bound_all} (b), while the other two methods give no clear trend and contain confusing patterns. In order to evaluate the efficiency of these estimating methods, we apply the correlation analysis between two variables~\cite{Johnson2001}, i.e.,
the actual risks and these three estimated generalization errors. These evaluations are based on 325 classifiers as in Fig.~\ref{gen_err_bound_all} (a), and the statistical r-values of them are 0.805, 0.372, and -0.230 as shown in Fig.~\ref{gen_err_bound_all} (b-d), respectively. 
The r-values also confirm that CV Bound and actual risk have high correlation, while the other two methods give very low correlation.
They show that $k$-fold cross-validation is more suitable to be the measure for the performance of binary classifiers. 
According to the above reason, we apply this measure in our research.

\section{The proposed methods}\label{Proposed_Method}

The combination of binary SVMs with high generalization performance
directly affects the accuracy of the multi-class classification.
In this section, we introduce four enhanced approaches
based on the previous techniques i.e., the ADAG, the DDAG, and Max Wins
by applying the generalization abilities in order to select 
suitable binary classifiers. An improvement of the ADAG is called
the Reordering Adaptive Directed Acyclic Graph (RADAG). There are two improved versions for the DDAG
i.e., Strong Elimination of the classifiers (SE) and Weak Elimination of the classifiers (WE).
The last technique is Voting-based Candidate Filtering (VCF) enhanced from Max Wins.
To increase the classification accuracy, the generalization estimated by $k$-fold cross-validation is utilized as the goodness measure of classifiers in our frameworks.

\subsection{Reordering Adaptive Directed Acyclic Graph}\label{RADAG}

The ADAG is designed to reduce the number of times the binary classifiers 
related to the target class are applied, from at most $N-1$ times required by the DDAG, to  $\lceil log_2N \rceil$ times or less. 
However, binary classifiers in the first level of the ADAG are still randomly selected,
and its misclassification can be produced 
at the time even when only one BCRT gives a wrong answer.  
In this section, we introduce a more effective method which uses 
the minimum weight perfect matching to select the optimal pairs
of classes in each level with minimum generalization error.
We called the method the Reordering Adaptive Directed Acyclic Graph (RADAG).

The structure of the RADAG is similar to the ADAG, but they
are different in the initialization of the binary classifiers 
in the top level and the order of classes in lower levels (see Fig.~\ref{RADAG}).
The reordering algorithm with minimum weight perfect matching is described 
in Algorithm~\ref{RADAG_algo}. The algorithm selects the optimal order of classes
in each level. It is different from the ADAG in that the initial order of classes in the ADAG is obtained randomly, 
and the matching of classes in successive levels depends on 
the classification results of nodes from the previous level. 
For the RADAG, the reordering process will be applied to the remaining candidate classes in all levels for determining the optimal sequence of them. 

\begin{figure}[htbp]
\centering
\scalebox{0.7}{\includegraphics{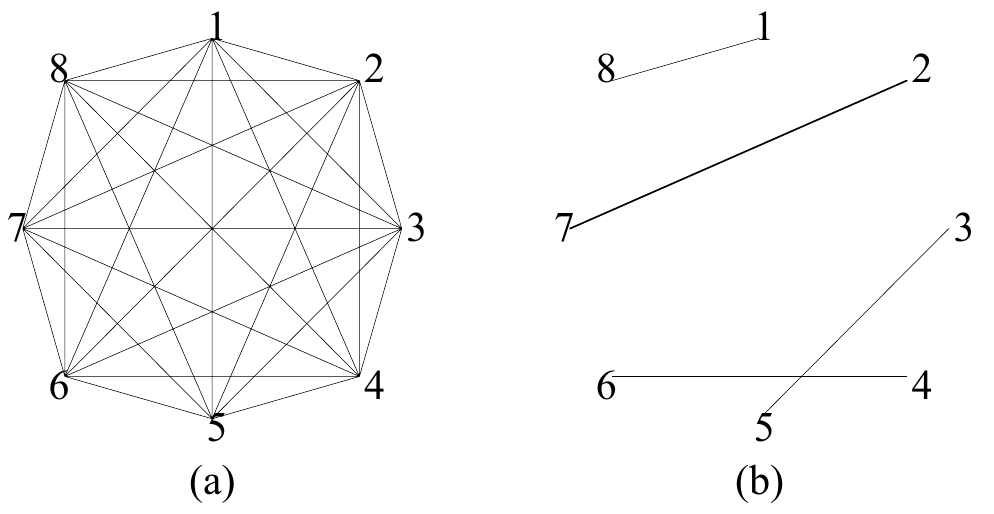}}
\caption{(a) A graph for an 8-class problem
(b) An example of the output of the reordering algorithm.}
\label{MWPM}
\end{figure}

To select the optimal set of classifiers, the generalization measure in Section~\ref{Estimate_Generr} 
is used as a criterion. This scheme provides less chance to predict the wrong class from all possible
$\frac{N!}{2^{\lfloor{N/2}\rfloor}{\lfloor{N/2}\rfloor}!}$ orders. Among $N(N-1)/2$ classifiers,
$N/2$ classifiers which have the smallest sum of generalization errors will be used in the classification.

\begin{algorithm*}[t]
  \caption{Reordering Adaptive Directed Acyclic Graph (RADAG).}\label{RADAG_algo}
  \begin{algorithmic}[1]
  \Procedure{RADAG}{}
  \State Initial set of candidate output classes $C=\{1, 2, 3, ..., N\}$, and set of discarded classes $D=\emptyset$
  \State Calculate generalization errors of all possible pairs of classes on $C$ as described in Section~\ref{Estimate_Generr}
  \State Create the binary SVMs from all possible pairs of classes on $C$
  \While{$|C|> 1$}
    \State Apply the minimum weight perfect matching~\cite{Cook97} to find the optimal 
$\lfloor \frac{|C|}{2} \rfloor$ pairs of classes from all possible 
\hspace*{3em}pairs on $C$ to obtain the optimal binary models with minimum generalization error
     \State $D$ $\gets$ Classify the example by the optimal binary models, and find the defeated classes
    \State $C$ $\gets$ $C-D$   
  \EndWhile\label{RADAG_Classify}
  \State {\it final output class} $\gets$ the last remaining candidate class 
  \State \textbf{return} {\it final output class}
  \EndProcedure
  \end{algorithmic}
\end{algorithm*}

Let $G = (V, E)$ be a graph with node set $V$ and edge set $E$. Each node in $G$
denotes one class and each edge indicates one binary classifier of which generalization error is estimated from Section~\ref{Estimate_Generr} (see Fig.~\ref{MWPM}(a)).
The output of the reordering algorithm for graph $G$ is a subset of edges with the
minimum sum of generalization errors of all edges and each node in $G$ is met by
exactly one edge in the subset (see Fig.~\ref{MWPM}(b)). 

Given a real weight $\epsilon_e$
being generalization error for each edge $e$ of $G$, the problem of reordering
algorithm can be solved by the minimum weight perfect matching~\cite{Cook97} that finds a perfect
matching $M$ of minimum weight $\sum(\epsilon_e : e \in M)$.

For $U \subseteq V$, let $E(U) = \{(i,j):(i,j) \in E, i \in U, j \in U\}$.
$E(U)$ is the set of edges with both endpoints in $U$. The set of edges incident to
node $i$ in the node-edge incidence matrix is denoted by $\delta(i)$. 
The convex hull
of perfect matchings on a graph $G = (V, E)$ with $|V|$ even is given by
\\ \hspace*{2em}a) $x \in \{0,1\}^m$
\\ \hspace*{2em}b) $\sum_{e \in \delta(v)}x_e = 1$ for $v \in V$
\\ \hspace*{2em}c) $\sum_{e \in E(U)}x_e \leq \lfloor \frac{|U|}{2} \rfloor$ for all odd sets $U \subseteq V$ with $|U| \geq 3$ or by (a),(b) and
\\ \hspace*{2em}d) $\sum_{e \in \delta(U)}x_e \geq 1$ for all odd sets $U \subseteq V$ with $|U| \geq 3$
\\where $|E| = m$, and 
$x_e = 1$ ($x_e = 0$) means that $e$ is (is not) in the matching.

Hence, the minimum weight
of a perfect matching is at least as large as the value of
\begin{equation}
min \sum_{e \in E} \epsilon_e x_e
\label{eqMWPM}
\end{equation}
\noindent where $x$ satisfies  \textquotedblleft (a), (b), and (c)\textquotedblright \, or \textquotedblleft (a), (b) and, (d)\textquotedblright. Therefore, the reordering problem can be solved by the integer program in Eq.~(\ref{eqMWPM}).

\subsection{Strong \& Weak Elimination of Classifiers for Enhancing Decision Directed Acyclic Graph}\label{EC_DDAG}

\begin{algorithm*}[t]
  \caption{Strong Elimination of the classifiers (SE).}\label{SE_algo}
  \begin{algorithmic}[1]
  \Procedure{SE}{}
  \State Initial set of candidate output classes $C=\{1, 2, 3, ..., N\}$, and set of discarded classes $D=\emptyset$
  \State Calculate generalization errors of all possible pairs of classes on $C$ as described in Section~\ref{Estimate_Generr}
  \State Create the binary models from all possible pairs of classes on $C$
  \State Sort the list of the binary models in ascending order by the generalization errors 
  \State {\it current classifier} $\gets$ the first element on the sorted list 
  \While{$|C|> 1$}  
    \State $D$ $\gets$ Classify the example by the current classifier, and find the defeated class
    \State $C$ $\gets$ $C-D$   
    \State {\it current classifier} $\gets$ the next element on the sorted list where it is not related to any classes discarded 
											\hspace*{12.36em}from $C$   
  \EndWhile\label{SE_Classify}
  \State {\it final output class} $\gets$ the last remaining candidate class 
  \State \textbf{return} {\it final output class}
  \EndProcedure
  \end{algorithmic}
\end{algorithm*}

\begin{algorithm*}[t]
  \caption{Weak Elimination of the classifiers (WE).}\label{WE_algo}
  \begin{algorithmic}[1]
  \Procedure{WE}{}
  \State Initial set of candidate output classes $C=\{1, 2, 3, ..., N\}$, and set of discarded classes $D=\emptyset$
  \State Calculate generalization errors of all possible pairs of classes on $C$ as elaborated in section \ref{Estimate_Generr}
  \State Create the binary models from all possible pairs of classes on $C$
  \State Sort the list of the binary models in ascending order by the generalization errors 
  \State {\it current classifier} $\gets$ the first element on the sorted list 
  \While{$|C|> 1$}  
    \State $D$ $\gets$ Classify the example by {\it current classifier}, and find the defeated class
    \State $C$ $\gets$ $C-D$   
    \State {\it current classifier} $\gets$ the next element on the sorted list where it does not include all two classes discarded
												\hspace*{11.8em} from $C$    
			 
  \EndWhile\label{WE_Classify}
  \State {\it final output class} $\gets$ the last remaining candidate class 
  \State \textbf{return} {\it final output class}
  \EndProcedure
  \end{algorithmic}
\end{algorithm*}

According to the characteristic of the DDAG, binary classification results 
of the previously employed binary classifiers are used to eliminate 
the candidate output classes, and thus the final output class will be assigned with one of the remaining candidate classes. 
By using the random technique for selecting a binary classifier, the DDAG produces mis-classification at the time when a BCRT with very low performance 
is selected and provides the wrong answer, 
as the target class will be discarded from the remaining candidate classes,
and it is not possible to reach the correct output class.  
In this section, we propose the framework to enhance the performance of the DDAG
to select the binary classifier with high performance
based on the generalization abilities of binary classifiers
as described in Section~\ref{Estimate_Generr}.

\begin{figure*}[htbp]
\centering
\vspace{-25pt}
\scalebox{1.2}{\includegraphics{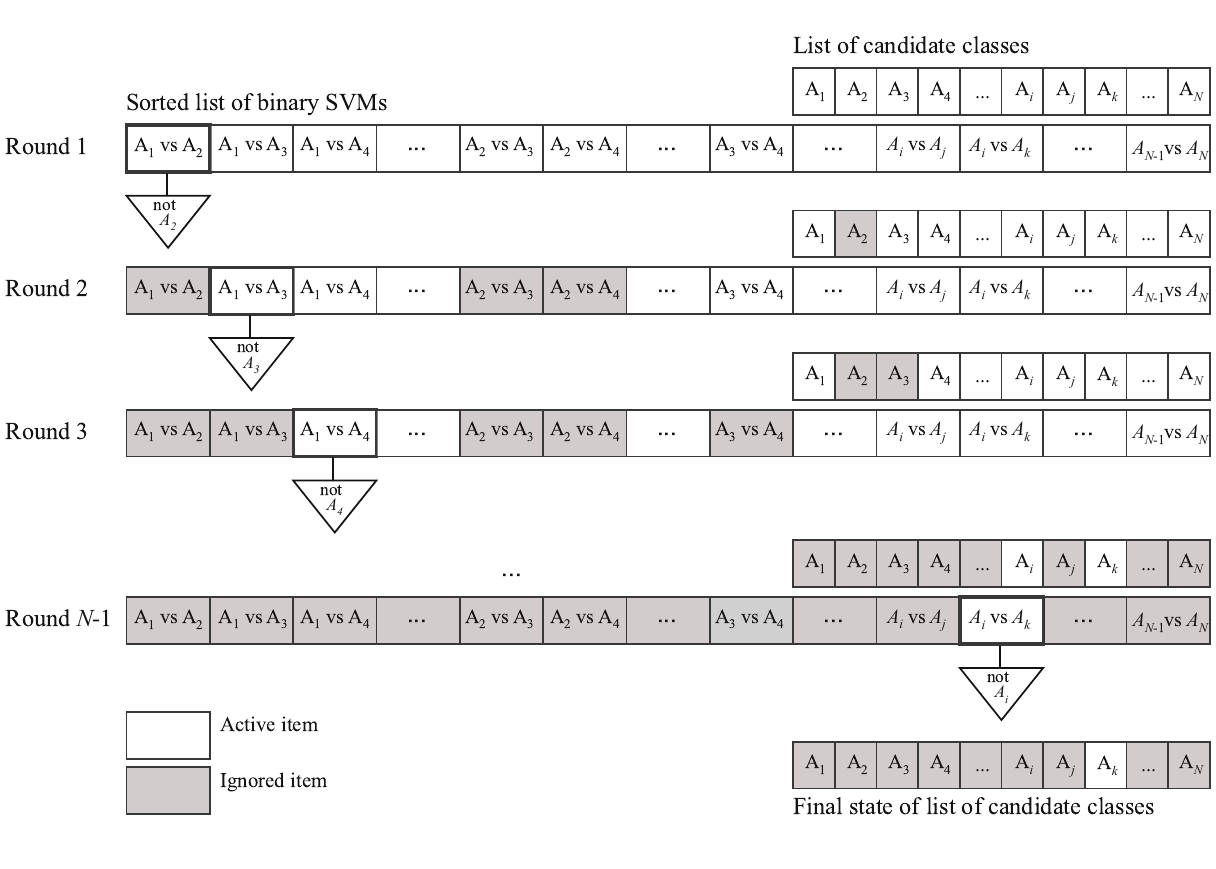}}
\vspace{-18pt}
\caption{Classification process of SE for an $N$-class problem.} 
\label{SE_structure}
\end{figure*}

\begin{figure*}[htbp]
\centering
\vspace{-20pt}
\scalebox{1.2}{\includegraphics{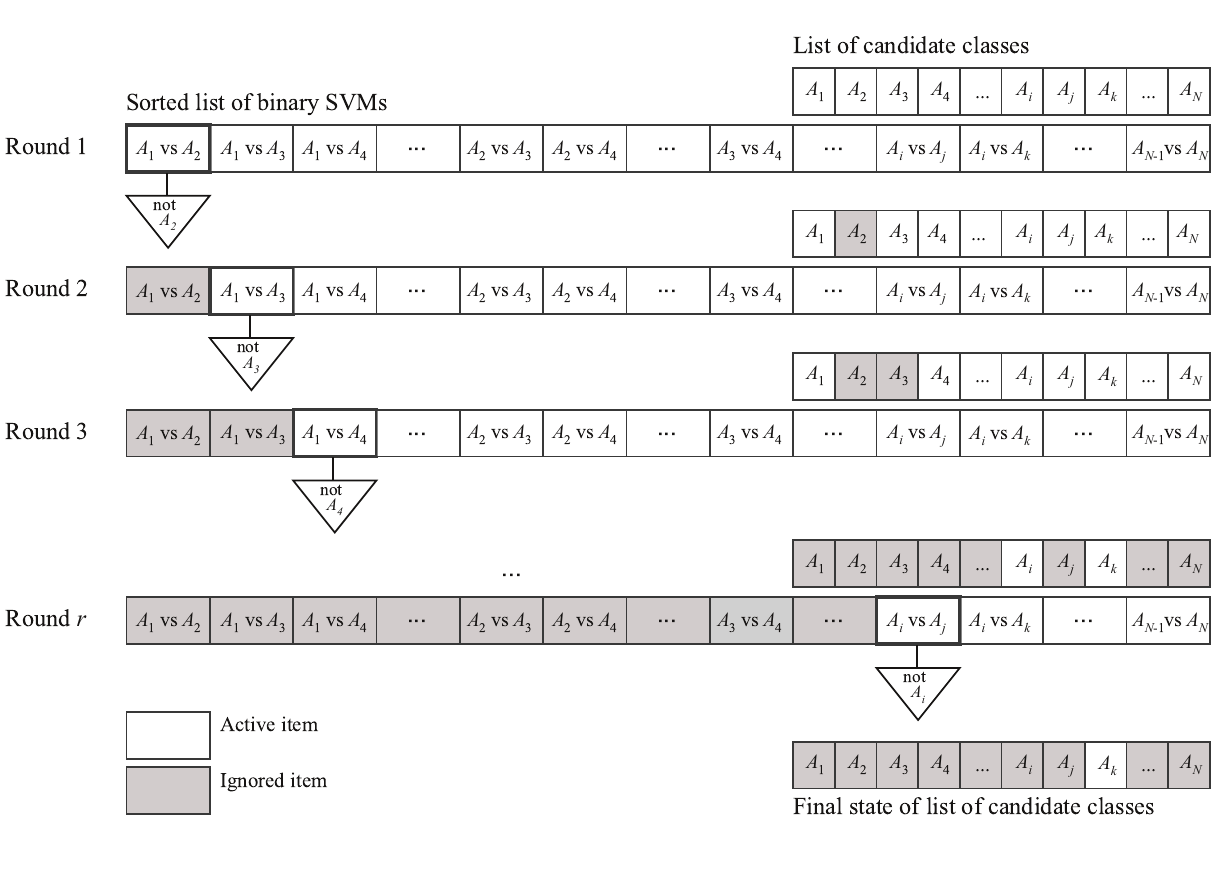}}
\vspace{-18pt}
\caption{Classification process of WE for an $N$-class problem.} 
\label{WE_structure}
\end{figure*}

We propose two methods that are Strong Elimination of the classifiers (SE)
and Weak Elimination of the classifiers (WE). Both algorithms 
are described in Algorithm~\ref{SE_algo} and Algorithm~\ref{WE_algo}.    
We also show a classification process of SE and WE for an $N$-class problem 
in Fig.~\ref{SE_structure} and Fig.~\ref{WE_structure}, respectively.

For both of the DDAG and SE, in each round, a defeated class will be 
removed from candidate output classes, and all binary classifiers related to
the defeated class are ignored. Due to this reason,  
they guarantee $N-1$ number of classifications for an $N$-class problem.
However, these ignored classifiers may have high generalization abilities and thus are helpful to eliminate the other remaining candidate classes. 
Therefore, we then propose WE to make use of binary classifiers with high generalization abilities.

According to the classifier elimination of WE, the number of classifications is bounded 
with the best case of $N-1$, and the worst case of $N(N-1)/2$. 
However, WE provides the opportunities to employ better classifiers
as shown in the Fig.~\ref{WE_structure}. At round $r$, suppose that classifier $A_i$ vs $A_j$ 
has lower generalization error than classifier $A_i$ vs $A_k$ , and both of them
are active classifiers. In this case, it is possible that classifier $A_i$ vs $A_j$ 
can remove the class $A_i$ from the list of two remaining candidate classes, and can avoid
using classifier $A_i$ vs $A_k$ with lower reliability that is unavoidable for SE as shown in  Fig.~\ref{SE_structure}.

\subsection{Voting Based Candidate Filtering}\label{VCF}

\begin{figure*}
        \centering
        \begin{subfigure}[b]{0.42\textwidth}      
				\centering                          
                \includegraphics[width=\textwidth]{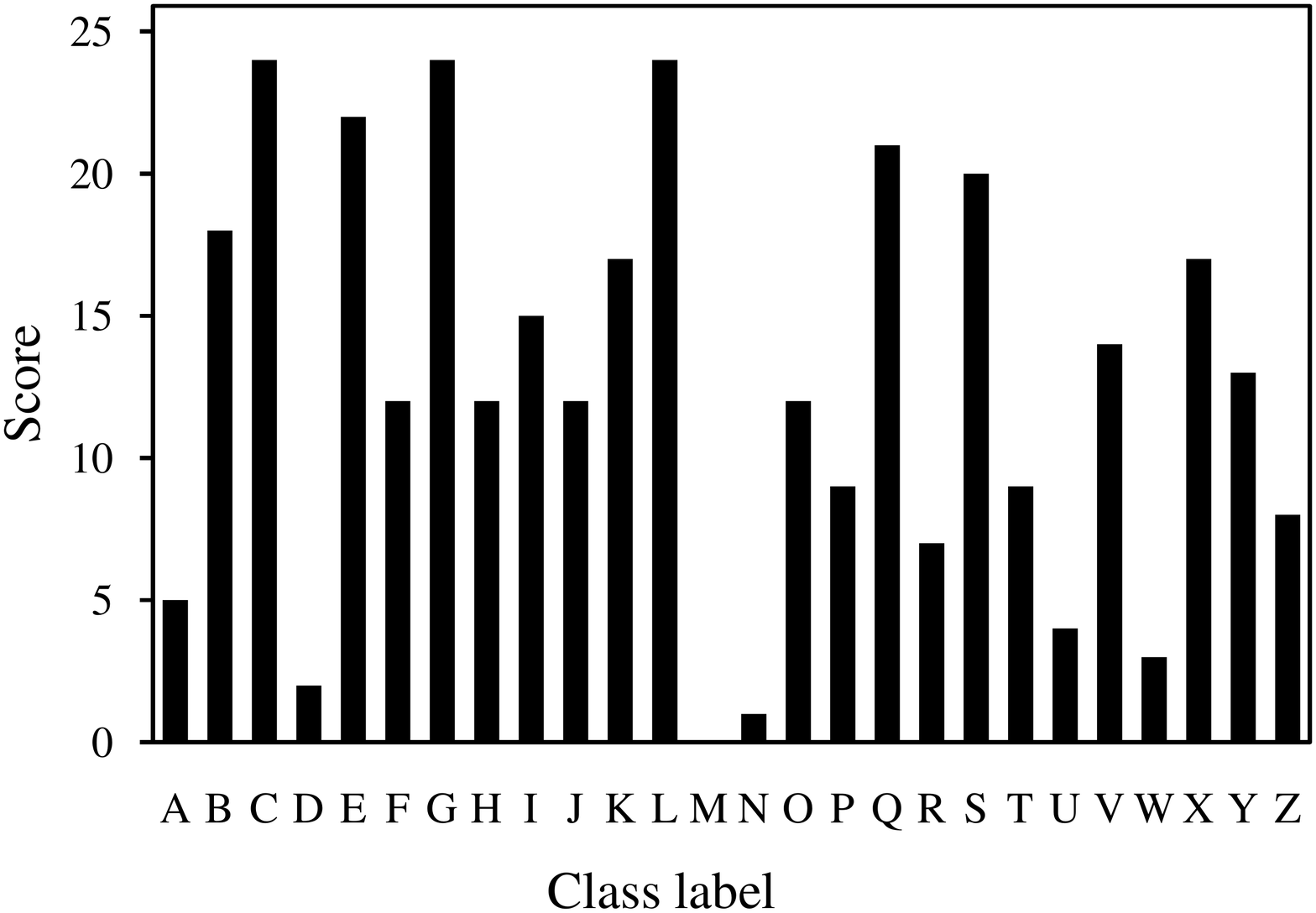}
                \vspace{-20pt}
                \caption{}
                \label{fig:fig09a}
        \end{subfigure}%
        \begin{subfigure}[b]{0.42\textwidth}
				\centering
                \includegraphics[width=\textwidth]{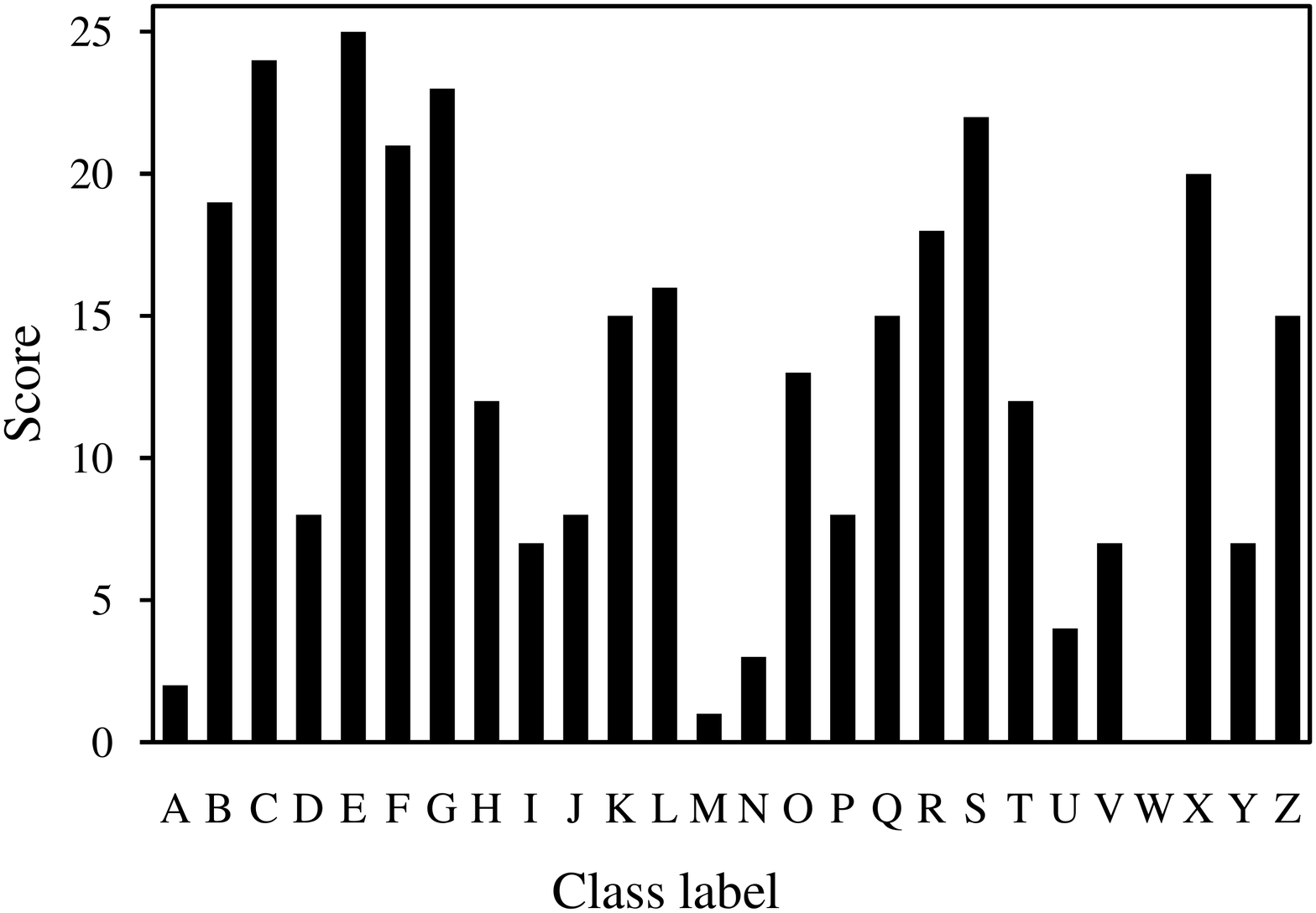}
                \vspace{-20pt}
                \caption{}
                \label{fig:fig09b}
        \end{subfigure}     		
        \caption{An example of high risk of misclassification of Max Wins together with score distribution of all classes: 
two cases of misclassification of class $\textquoteleft$C' due to only one BCRT giving the wrong answer 
in the Letter problem having 26 classes (25 possible BCRTs, and 25 points as the largest possible score), 
a) three classes, including $\textquoteleft$C', $\textquoteleft$G', and $\textquoteleft$L', with 
equal score (only one BCRT $\textquoteleft$C vs G' giving the wrong class), 
and b) the non-target class $\textquoteleft$E' with the highest score (only one BCRT $\textquoteleft$C vs E' providing the wrong class).}
\label{let_maxwin_score}
\end{figure*}

Max Wins is one of high performance techniques that work based on the concept of {\it``trust on the most popular opinion''} 
for making decision to select the output class. If all of $N-1$ BCRTs
give the correct answer, Max Wins will always provide the correct output class. 
It does not depend on the answers of the other binary classifiers.
However, if only one of BCRTs gives a wrong answer, it may lead to misclassification due to equal voting, or another non-target class
reaching the largest vote. Fig.~\ref{let_maxwin_score} shows an example of such cases, taken from our experiment on the Letter dataset (see Section~\label{experiments} for more details); Fig.~\ref{let_maxwin_score}(a) and (b) show the cases of equal voting and another non-target class having the largest vote, respectively.

We propose a novel multi-class classification approach that alleviates the above problem of Max Wins, and uses the same concept 
{\it``trust on the most popular opinion''} for filtering out the low competitive classes.
On the other hand, high competitive classes will be voted to be candidate output classes, 
though there exist some BCRTs providing the wrong answer.
If there is more than one remaining class, 
the output class will be selected via the mechanism of WE.
Our proposed technique aims to combine the strong point of both Max Wins and WE, 
and is called  Voting based Candidate Filtering (VCF). The details of our algorithm are shown 
in Algorithm~\ref{VCF_algo}.

\begin{algorithm*}[t]
  \caption{Voting based Candidate Filtering (VCF).}\label{VCF_algo}
  \begin{algorithmic}[1]
  \Procedure{VCF}{}
  \State Initial set of candidate output classes $C=\{1, 2, 3, ..., N\}$, and score of class $i$: $s_{i \in N} \gets 0$
  \State Create the binary models from all possible pairs of classes on $C$
\For{$j$=1 to $N(N-1)/2$ }  
  \State $w$ $\gets$ Classify the example by classifier $j^{th}$, and find the winner class 
  \State $s_w$ $\gets$ $s_w+1$
\EndFor   
\State $s_{top}$ $\gets$ Find the top voting score of all $s_{i \in N}$
\For{$i$=1 to $N$ }
    \State $dp_i \gets \frac{(s_{top}-s_i) \times 100}{s_{top}}$  
    \If {$dp_i \leq$ {\it $threshold\_value$}}
     \State Add class $i$ into the set of candidate output classes $C$ 
    \EndIf     
  \EndFor    
  \If {$|C|> 1$}
     \State {\it final output class} $\gets$ Call the WE procedure   
  \Else
     \State {\it final output class} $\gets$ the last remaining candidate class 
  \EndIf
  \State \textbf{return} {\it final output class}
  \EndProcedure
  \end{algorithmic}
\end{algorithm*}

Let  $s_{top}$, and $s_i$ indicate the maximum of scores for all $N$ classes, and the score of class $i \in [N]$ for a test data, respectively. 
Also let $dp_i$ denotes the percentage of the difference between $s_{top}$ and $s_i$.
An example of the calculation of $dp_i$ is shown in Fig.~\ref{let_maxwin_score} (a),
where $i= \textquoteleft$E', the score of class $\textquoteleft$E' = 23 points, 
and the score of class $\textquoteleft$C' = 24 points (as the top score). Then
$dp_i$ value can be calculated by $\frac{(24 - 23)\times 100}{24}=4.17$.
We also define {\it $threshold\_value$} to be the threshold of $dp_i$ for considering class $i$ as a candidate for the target class;
class $i$ will be accepted into the set of high competitive candidate classes 
if and only if its $dp_i$ is less than or equal to {\it $threshold\_value$}.
We want to keep the size of the filtered candidate classes as small as possible while still containing the target class. 

\begin{figure*}[htbp]
\centering
\scalebox{0.33}{\includegraphics{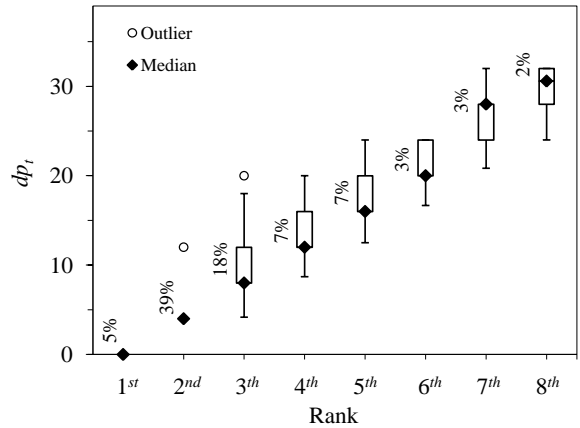}}
\vspace{-20pt}
\caption{
A case study of the $461$ examples with {\it high risk} of misclassification in the Letter problem. 
The maximum voting scores of these examples are reached (1) by both of the target class and the non$-$target class (equal vote: $1^{st}$ rank), 
or (2) by a non$-$target class (absolutely wrong: $2^{nd}-8^{th}$ rank).  
The figure shows the target class score of these examples by observing between the $dp_t$ and the rank of the target class. 
}
\label{maxwin_risk}
\end{figure*}

A case study of high risk of misclassification in the Letter dataset including 4,010 
examples where Max Wins provides 3,549 examples with the correct result, and 461 examples with high risk of misclassification. By a high-risk example, we mean (1) the example with an equal vote (the score of the target class is equal to those of other non-target classes) and (2) the example with a vote less than the maximum vote that is then mis-classified by Max Wins. 
These high risk examples will be hopefully recovered with the correct class label by our proposed algorithm.
In our experiment, the high-risk examples includes 24 examples (around 5$\%$) with an equal vote, and 437 examples (around 95$\%$) with a vote less than the maximum as shown in Fig.~\ref{maxwin_risk},  
where $dp_t$ represents the percentage of the difference between $s_{top}$ and the score of the target class.
For each example, we calculate the rank of the voting score of the target class compared to the other non-target classes, and consider only the first eight ranks.
There are 24 examples (around $5\%$) in the first rank, while in the second to the eighth ranks, 
the numbers of examples are 171, 77, 31, 30, 15, 14, and 10 (around $39\%, 18\%, 7\%, 7\%, 3\%, 3\%$, and $2\%$), respectively.
The examples with the different ranks have different ranges of $dp_t$ values, 
such as,
in the second rank, the $dp_t$ values are varied from $4.0$ to $12.0$,  
in the third rank, the $dp_t$ values are varied from $4.2$ to $20.0$, 
in the fourth rank, the $dp_t$ values are varied from $8.7$ to $20.0$, 
and so on.

According to this case study, there can be at most $5\%$ of examples that will be correctly classified with the correct class label by random selection of Max Wins, while the other $95\%$ of examples will be absolutely misclassified.
We want to recover an example that is not correctly classified by Max Wins, as its actual target class is not in the first rank  
or its target class has equal vote with some other output classes. If {\it $threshold\_value$} is set as $1$ in the VCF algorithm, it will guarantee that all high-risk misclassified examples with $dp_t$ values 
no greater than can be filtered into the set of the candidate output classes; 
in this case only the examples in the first rank ($5\%$ of examples) will be selected.
When we apply a bigger threshold, e.g. {\it $threshold\_value$} $=10$, 
it covers all misclassified examples in the first and the second ranks ($5\%+39\%$), 
almost of the third rank ($18\%$), and some parts of the fourth rank ($7\%$).
It shows that the increase of {\it $threshold\_value$} covers more candidate classes,
while the larger size of {\it $threshold\_value$} 
creates a higher risk to employ an unnecessarily large number of binary classifiers.
On the other hand, if {\it $threshold\_value$} is too low, the target class may be removed.
However, a suitable {\it $threshold\_value$} can be obtained by general tuning techniques.
For our experiment, we just define {\it $threshold\_value$} to be $10$ for all of datasets without fine-tuning which is good enough to demonstrate the effectiveness of the VCF algorithm.

\section{Experiments}\label{experiments}

In this section, we design the experimental setting to evaluate the performance 
of the proposed methods. We compare our methods with the traditional algorithms, i.e., the DDAG, the ADAG, and Max Wins.
We divide this section into two parts as experimental protocols, and results \& discussions.

\vspace*{-0.5em}\subsection{Experimental Protocol}\label{ExpProtocol}
We run experiments on sixteen datasets from the UCI Machine Learning
Repository~\cite{Blake98} including {\it Page Block, Glass, Segment, Arrhyth, 
Mfeat-factor, Mfeat-fourier,
Mfeat-karhunen, Mfeat-zernike, Optdigit, Pendigit, Primary tumor, Libras Movement, Abalone, Krkopt, Spectrometer}, 
and {\it Letter} (see Table~\ref{Dataset_Dessciption}). For the datasets containing both training data and test data, 
we added up both of them into one set, and used 5-fold cross validation for evaluating the classification accuracy.

\begin{table}[htp]
\centering
\caption[]{{\footnotesize Description of the datasets used in the experiments.}}
{\scriptsize
\begin{tabular}{l r r r}
\hline
\textbf{Datasets} & \textbf{\#Cases} & \textbf{\#Classes} & \textbf{\#Features}\\
\hline
Page Block & 5,473 & 5 & 10 \\
Glass & 214 & 6 & 9 \\
Segment & 2,310 & 7 & 18 \\
Arrhyth	&438&	9&	255\\
Mfeat-factor&2,000&10&216\\
Mfeat-fourier&2,000&10&76\\
Mfeat-karhunen&2,000&10&64\\
Mfeat-zernike&2,000&10&47\\
Optdigit&5,620&10&62\\
Pendigit&10,992&10&16\\
Primary tumor&315&13&15\\
Libras Movement&360&15&90\\
Abalone&4,098&16&8\\
Krkopt&28,056&18&6\\
Spectrometer&475&21&101\\
Letter & 20,052 & 26 & 16 \\
\cline{1-4}
\end{tabular}}
\label{Dataset_Dessciption}
\end{table}

In these experiments, we scaled data to be in [-1,1] and employed two kernel functions
i.e., the Polynomial kernel $K(x_i,x_j) \equiv |({\bf x_i}\cdot{\bf x_j}+1)|^d$, 
and the RBF kernel $K(x_i,x_j) \equiv e^{-\gamma ||{\bf x_i}-{\bf x_j}||^2}$.
For the polynomial kernel we applied the same set of degrees $d=\{2,3,4,5\}$ to all datasets,
and for the RBF kernel we applied the set of degrees $\gamma_1=\{1,0.5,0.1,0.05\}$ to {\it Page Block,  Glass, Segment, Mfeat-zernike, 
Pendigit, Libras Movement, Abalone, Krkopt}, and {\it Letter}, and applied the set of degrees $\gamma_2=\{0.1,0.05,0.01,\\0.005\}$
to the other datasets. The default parameter of regularization parameter $C$ was used for model construction; this parameter is used to trade off between error of the SVM on training data and margin maximization.
In the training phase, we used software package $SVM^{light}$ version 6.02~\cite{Joachims98,Joachims99}
to create the $N(N-1)/2$ binary classifiers. For the DDAG and the ADAG, we examined all possible orders of classes
for datasets having not more than 8 classes, whereas we randomly selected 50,000 orders for datasets
having more than 8 classes, and we then calculated the average of accuracy of these orders.

\vspace*{-0.5em}\subsection{Results \& Discussions}\label{ExpResult}

We compare the original methods with their enhanced techniques in three tasks including: 
(1) the ADAG with the RADAG, (2) the DDAG with two improved approaches, i.e., SE and WE, 
and (3) Max Wins with VCF.

We also selected the best techniques from (1) and (2), i.e., the RADAG and WE, respectively, 
and compared them with Max Wins as the state of the art technique. 
These comparison results are shown in Table~\ref{ADAG_RADAG_comparison} to Table~\ref{Max_wins_RADAG_WE_comparison}. 
Moreover, paired comparison among all of three traditional methods (the DDAG, the ADAG, and Max Wins), 
and all proposed techniques (SE, the RADAG, WE, and VCF) are concluded in Table~\ref{pair_comparison}.

The best accuracy among these methods is represented in bold-face.
In addition, we used the one-tailed paired t-test technique to analyze
the significant difference between the accuracies of the traditional
algorithms and the proposed algorithms. To estimate the difference
between accuracies, we use a $k$-fold cross-validation method~\cite{Mitchell97}.

To indicate the level of the confidence interval using a one-tailed paired t-test
in the Table~\ref{ADAG_RADAG_comparison} to 
Table~\ref{Max_wins_RADAG_WE_comparison},
the symbol \textquoteleft$+$' and \textquoteleft$-$'
are used to represent that the corresponding method has higher accuracy, and lower accuracy
compared to a baseline method, respectively. The number of symbols shows the level of confidence interval
for estimating the difference between accuracies of two algorithms i.e., one symbol, two symbols, and three symbols represent 90\%, 95\%, and 99\% respectively.

\begin{table*}[htp]
\centering
\caption[]{{\footnotesize A comparison of the classification accuracy of the ADAG and the RADAG. }}
{\scriptsize
\begin{tabular}{l l l l l l }
\hline
{} & \multicolumn{2}{c}{Polynomial}& \hspace*{1em} &\multicolumn{2}{c}{RBF}\\
\cline{2-3} \cline{5-6}
{Data sets}&{ADAG\hspace*{1em} }&\multicolumn{1}{c}{RADAG}&{}& {ADAG\hspace*{1em} }&\multicolumn{1}{c}{RADAG}\\
\hline
Page Block      &\textbf{93.597}&93.541\hspace*{0.36em}$---$              			  	&{ }&\textbf{93.562}&93.555\hspace*{0.36em}$-$\\
Glass        	&63.879&\textbf{64.019}                               					&{ }&63.084&\textbf{63.318}\hspace*{0.36em}\\
Segment         &93.207&\textbf{93.236}$+{\color{white}*}$           					&{ }&93.348&\textbf{93.366}\\
Arrhyth         &\textbf{63.489}&63.470                               					&{ }&\textbf{58.049}&57.991\hspace*{0.36em}\\
Mfeat-factor    &\textbf{97.238}&97.225\hspace*{0.36em}$--{\color{white}*}$           	&{ }&96.921&\textbf{96.938}\\
Mfeat-fourier   &82.839&\textbf{82.863}                               					&{ }&82.456&\textbf{82.513}$+$\\
Mfeat-karhunen  &\textbf{96.864}&96.863           										&{ }&96.890&\textbf{96.900}\\
Mfeat-zernike   &82.368&\textbf{82.413}$+++{\color{white}*}$            				&{ }&81.867&\textbf{81.888}\\
Optdigit        &98.995&\textbf{98.999}                              					&{ }&98.620&\textbf{98.630}\\
Pendigit        &99.400&\textbf{99.402}                               					&{ }&99.313&\textbf{99.320}$+$\\
Primary tumor   &47.266&\textbf{47.619}$++{\color{white}*}$           					&{ }&46.089&\textbf{46.429}$+++$\\
Libras Movement          &\textbf{73.218}&73.194                               					&{ }&72.289&\textbf{72.569}$+$\\
Abalone         &27.603&\textbf{27.648}                               					&{ }&\textbf{27.353}&27.337\\
Krkopt          &53.102&\textbf{53.239}$+++$        			      					&{ }&53.088&\textbf{53.173}$+++$\\
Spectrometer    &54.445&\textbf{54.842}$++{\color{white}*}$           					&{ }&50.808&\textbf{51.579}$+{\color{white}*}$\\
Letter          &88.668&\textbf{88.787}$+++$        				  					&{ }&89.989&\textbf{90.090}$+++$\\
\hline
\end{tabular}}
\label{ADAG_RADAG_comparison}
\end{table*}

The experimental results in Table~\ref{ADAG_RADAG_comparison}
uses the ADAG as the baseline algorithm.
It shows that the RADAG yields highest accuracy in several datasets.
The results also show that, at $95$\% confidence interval, the RADAG performs
statistically better than the ADAG in five datasets using the Polynomial kernel
and better in three datasets using the RBF kernel.
As shown in the table, the RADAG performs better when the number of classes is comparatively large, 
and does not perform well in the datasets with the small number of classes, i.e., the Page Block, and the Mfeat-factor with 5 and 10 classes, respectively.
We believe that in case of datasets with the large number of classes, the variety of generalization errors of classifiers in consideration is rich and the RADAG is able to choose good classifiers freely, whereas the RADAG may be forced to select ineffective classifiers in case of the small number of classes, and it could lead to an incorrect output class.  

\begin{table*}[htp]
\centering
\caption[]{{\footnotesize A comparison of the classification accuracy between the DDAG, and our methods, i.e, SE, and WE. }}
{\scriptsize
\begin{tabular}{l c c l l l c c l l }
\hline
&\multicolumn{4}{c}{Polynomial}& \hspace*{1em} &\multicolumn{4}{c}{RBF}\\
\cline{2-5} \cline{7-10}
{Data sets}&\multicolumn{2}{l}{DDAG}&\multicolumn{1}{c}{ SE }&\multicolumn{1}{c}{WE}&
&\multicolumn{2}{l}{DDAG}&\multicolumn{1}{c}{ SE }&\multicolumn{1}{c}{WE}\\
\hline
Page Block      &93.597&\hspace*{1em}   &93.541\hspace*{0.36em}$---$            			&\textbf{93.623}$+$&
                &93.562&\hspace*{1em}   &93.555\hspace*{0.36em}$-$                      	&\textbf{93.582}$++{\color{white}*}$\\
Glass        	&63.892&                &\textbf{64.019}                             		&\textbf{64.019}
                &&63.084&               &\textbf{63.201}            						&\textbf{63.201}\\                
Segment         &93.207&                &93.236\hspace*{0.36em}$+$                    	&\textbf{93.247}$++$
                &&93.350&               &93.344                             			&\textbf{93.366}\\
Arrhyth         &63.490&                &\textbf{63.527}                   				&\textbf{63.527}
                &&58.048&               &57.991\hspace*{0.36em}                      	&\textbf{58.162}\\
Mfeat-factor    &97.238&                &\textbf{97.250}                    			&97.238
                &&96.923&               &\textbf{96.975}$+$           					&\textbf{96.975}$++{\color{white}*}$\\
Mfeat-fourier   &82.837&                &\textbf{82.863}                    			&\textbf{82.863}
                &&82.443&               &82.475                             			&\textbf{82.538}\\
Mfeat-karhunen  &96.863&                &\textbf{96.875}                    			&\textbf{96.875}
                &&96.861&               &96.850           								&\textbf{96.988}$+{\color{white}*}$\\               
Mfeat-zernike   &82.362&                &\textbf{82.400}$++{\color{white}*}$           &82.350
                &&81.869&               &\textbf{81.888}         						&81.863\\
Optdigit        &98.994&                &\textbf{99.013}$++{\color{white}*}$    		&99.008\hspace*{0.36em}$+{\color{white}*}$
                &&98.618&               &\textbf{98.643}$+$           					&98.630\\
Pendigit        &99.399&                &\textbf{99.404}                    			&99.402
                &&99.312&               &99.318           								&\textbf{99.320}$+$\\
Primary tumor   &47.227&                &47.064                             			&\textbf{47.460}$++$
                &&46.019&               &46.032                             			&\textbf{46.111}\\
Libras Movement          &73.142&                &73.264                             			&\textbf{73.472}$++{\color{white}*}$
                &&72.283&               &\textbf{72.569}$+$           					&72.431\\
Abalone         &27.611&                &27.648\hspace*{0.36em}                    	&\textbf{27.672}$+$
                &&27.354&               &27.330                             			&\textbf{27.398}\\
Krkopt          &53.101&                &53.263\hspace*{0.36em}$++$      				&\textbf{53.472}$+++$
                &&53.088&               &53.212\hspace*{0.36em}$+++$      				&\textbf{53.320}$+++$\\
Spectrometer    &54.373&                &\textbf{54.632}           					&54.421
                &&50.821&               &51.316                             			&\textbf{51.842}$++$\\
Letter          &88.609&                &88.707\hspace*{0.36em}$+++$\hspace*{0.6em}    	&\textbf{88.835}$+++$
                &&89.903&               &89.977\hspace*{0.36em}$++$\hspace*{0.6em}   	&\textbf{90.294}$+++$\\
\hline
\end{tabular}}
\label{DDAG_SE_WE_comparison}
\end{table*}

Table~\ref{DDAG_SE_WE_comparison} shows the experimental results of SE and WE compared with
the DDAG as the baseline algorithm.
Both WE and SE have higher accuracy than the traditional DDAG
in almost all datasets. The results also show
that at $95$\% confidence interval, SE performs statistically significantly better
than the DDAG in four datasets using the Polynomial kernel
and significantly better than the DDAG in two datasets using the RBF kernel.
It is similar to the previous comparison between the ADAG and the RADAG that
in datasets with the small number of classes, the classifier manipulation of 
SE may be forced to select inaccurate classifiers and it possibly leads to the misclassification.
The results also show that WE performs statistically significantly better
than the DDAG in five datasets in both cases of the Polynomial kernel
and the RBF kernel. 
These results illustrate that WE can reduce the risk of selecting inaccurate classifiers
compared to SE. 

\begin{figure}[htbp]
\vspace{-20pt}
\hspace{-20pt}
\scalebox{0.36}{\includegraphics{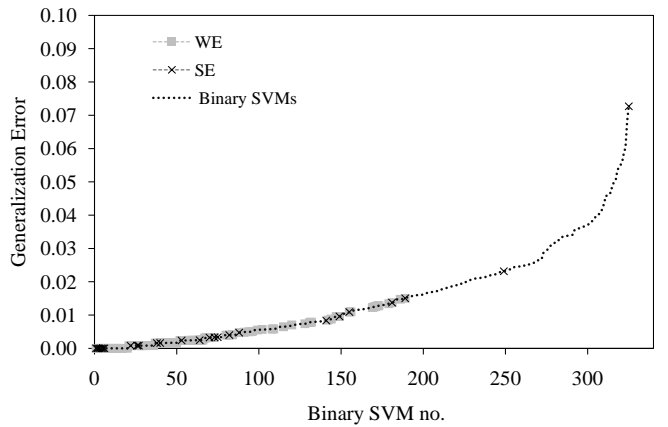}}
\vspace{-50pt}
\caption{An example of generalization errors of binary SVMs used by WE and SE in the Letter dataset.}
\label{example_of_WE_SE_in_Letter}
\end{figure}

\begin{table*}[htp]
\centering
\caption[]{{\footnotesize A comparison of the classification accuracy of Max Wins and VCF. }}
{\scriptsize
\begin{tabular}{l c l  l c l }
\hline
{Data sets} & \multicolumn{2}{c}{Polynomial}& \hspace*{1em} &\multicolumn{2}{c}{RBF}\\
\cline{2-3} \cline{5-6}
{}&\multicolumn{1}{c}{Max Wins}&\multicolumn{1}{c}{VCF}&{}& \multicolumn{1}{c}{Max wins}&\multicolumn{1}{c}{VCF}\\
\hline
Page Block        &93.600&\textbf{93.623}                    		&{}&93.567&\textbf{93.582}$++{\color{white}*}$\\
Glass          	  &63.863&\textbf{64.019}                   		    &{}&63.143&\textbf{63.201}\\
Segment           &93.209&\textbf{93.247}$++{\color{white}*}$      		&{}&93.351&\textbf{93.366}\\
Arrhyth           &63.489&\textbf{63.527}&{}                          	&58.048&\textbf{58.162}\\
Mfeat-factor      &\textbf{97.242}&97.238                             	&{}&96.927&\textbf{96.975}$++{\color{white}*}$\\
Mfeat-fourier     &\textbf{82.852}&82.825                             	&{}&82.454&\textbf{82.525}\\
Mfeat-karhunen    &\textbf{96.879}&96.875                             	&{}&96.952&\textbf{96.963}\\
Mfeat-zernike     &82.338&\textbf{82.350}                             	&{}&81.825&\textbf{81.863}\\
Optdigit          &99.004&\textbf{99.013}                             	&{}&\textbf{98.631}&98.630\\
Pendigit          &\textbf{99.402}&\textbf{99.402}            			&{}&99.315&\textbf{99.320}\\
Primary tumor     &47.394&\textbf{47.460}                    		  	&{}&\textbf{46.508}&46.191 \\
Libras Movement   &73.194&\textbf{73.472}$++{\color{white}*}$      		&{}&72.373&\textbf{72.431}\\
Abalone           &27.614&\textbf{27.672}$+$                    		&{}&27.375&\textbf{27.398}\\
Krkopt            &53.149&\textbf{53.475}$+++$      					&{}&53.146&\textbf{53.328}$+++$\\
Spectrometer      &54.263&\textbf{54.421}$+{\color{white}*}$&{}        &51.026&\textbf{51.842}$++$\\
Letter            &88.706&\textbf{88.869}$+++$      					&{}&90.112&\textbf{90.316}$+++$\\
\hline
\end{tabular}}
\label{Maxwin_VCF_comparison}
\end{table*}

\begin{table*}[htp]
\centering
\caption[]{{\footnotesize A comparison of the classification accuracy between Max Wins and the RADAG, and WE. }}
{\scriptsize
\begin{tabular}{l c l l l c l l }
\hline
&\multicolumn{3}{c}{Polynomial}& \hspace*{1em} &\multicolumn{3}{c}{RBF}\\
\cline{2-4} \cline{6-8}
{Data sets}&\multicolumn{1}{c}{Max Wins}&\multicolumn{1}{c}{ RADAG }&\multicolumn{1}{c}{WE}&
&\multicolumn{1}{c}{Max Wins}&\multicolumn{1}{c}{ RADAG }&\multicolumn{1}{c}{WE}\\
\hline
Page Block          &93.600         &93.541\hspace*{0.36em}$---$\hspace*{0.6em}     			&\textbf{93.623}
                    &&93.567        &93.555\hspace*{0.36em}$--{\color{white}*}$\hspace*{0.6em}  &\textbf{93.582}$++{\color{white}*}$\\
Glass            	&63.863         &\textbf{64.019}                                         	&\textbf{64.019}
                    &&63.143        &\textbf{63.318}                 							&63.201\\                    
Segment             &93.209         &93.236\hspace*{0.36em}$+{\color{white}*}$         		&\textbf{93.247}$++$
                    &&93.351        &\textbf{93.366}                                			&\textbf{93.366}\\
Arrhyth             &63.489         &63.470                                         			&\textbf{63.527}
                    &&58.048        &57.991\hspace*{0.36em}                                  &\textbf{58.162}\\
Mfeat-factor        &\textbf{97.242}&97.225\hspace*{0.36em}$---$                        		&97.238
                    &&96.927        &96.938                                         			&\textbf{96.975}$++{\color{white}*}$\\
Mfeat-fourier       &82.852         &\textbf{82.863}                                			&\textbf{82.863}
                    &&82.454        &82.513\hspace*{0.36em}                                	&\textbf{82.538}\\
Mfeat-karhunen      &\textbf{96.879} &96.863                                					&96.875
                    &&96.952        &96.900                                					    &\textbf{96.988}\\
Mfeat-zernike       &82.338 		&\textbf{82.413}$++{\color{white}*}$      					&82.350
                    &&81.825        &\textbf{81.888}$++{\color{white}*}$                     	&81.863\\
Optdigit            &99.004         &98.999                                         			&\textbf{99.008}
                    &&\textbf{98.631}&98.630                               						&98.630\\
Pendigit            &\textbf{99.402}&\textbf{99.402}                                			&\textbf{99.402}
                    &&99.315        &\textbf{99.320}                                			&\textbf{99.320}\\
Primary tumor       &47.394         &\textbf{47.619}$++$         								&47.460
                    &&\textbf{46.508}&46.429                                        			&46.111\\
Libras Movement     &73.194         &73.194                                         			&\textbf{73.472}$++{\color{white}*}$
                    &&72.373        &\textbf{72.569}                       					&72.431\\
Abalone             &27.614         &27.648                                         			&\textbf{27.672}$+$
                    &&27.375        &27.337                                         			&\textbf{27.398}\\
Krkopt              &53.149         &53.239\hspace*{0.36em}$++{\color{white}*}$             	&\textbf{53.472}$+++$
                    &&53.146        &53.173                                         			&\textbf{53.320}$+++$\\
Spectrometer        &54.263         &\textbf{54.842}$++$         								&54.421\hspace*{0.36em}$+$
                    &&51.026        &51.579                                         			&\textbf{51.842}$++$\\
Letter              &88.706         &88.787\hspace*{0.36em}$++{\color{white}*}$             	&\textbf{88.835}$+++$
                    &&90.112        &90.090                                         			&\textbf{90.294}$+++$\\
\hline
\end{tabular}}
\label{Max_wins_RADAG_WE_comparison}
\end{table*}

We further analyze the results comparing WE and SE on the Letter dataset
which consists of $26$ classes and 325 binary learners, as shown in Fig.~\ref{example_of_WE_SE_in_Letter}.
These 325 classifiers in the figure are sorted in ascending order by the generalization error, 
and this sequence of classifiers is maintained
in the classification phase. SE requires $25$ classifiers and WE requires $93$ classifiers in this case, and 
the generalization error of the worst binary classifier
in WE is almost five times lower than in SE 
(the largest generalization errors of all binary SVMs used in SE and WE
are 0.015, 0.073, respectively).
As a result, the average performance of the binary classifiers
in WE is higher than SE.

As shown in Table~\ref{Maxwin_VCF_comparison} with Max Wins as the baseline method, 
VCF yields higher accuracy than Max Wins in almost all of datasets.
The results show that, at $95$\% confidence interval, in the Polynomial kernel
VCF performs statistically significantly better than Max Wins in four datasets,
and in the RBF kernel VCF performs statistically significantly
better than Max Wins in five datasets.
The previous three tables show that our proposed methods improve
the accuracy of the ADAG, the DDAG, and Max Wins significantly.

Next, we select the best algorithm in each table from the first two tables, i.e, the RADAG, and WE, 
and then compare them to Max Wins.
According to experimental result in Table~\ref{Max_wins_RADAG_WE_comparison},
at $95$\% confidence interval, the RADAG performs statistically
significantly better than Max Wins in five datasets using the Polynomial kernel, and 
significantly higher than Max Wins in one dataset using the RBF kernel.
In case of the small number of classes, it is possible that the RADAG will have the effect mentioned above.
For WE, the results show that it performs statistically significantly better
than Max Wins in four datasets in case of the Polynomial kernel
and significantly better than Max Wins in five datasets in case of the RBF kernel.
There is no any dataset in which Max Wins has significantly higher accuracy than WE.

\begin{table*}[htp]
\centering
\caption[]{{\footnotesize Paired comparisons among all techniques including of three traditional techniques (DDAG, ADAG, and Max Wins), 
and four proposed techniques (RADAG, SE, WE, and VCF).}}
{\scriptsize
\begin{tabular}{l l r r r r r r r r }
\hline
\multicolumn{1}{c}{}&&\multicolumn{3}{c}{Traditional Methods}& \hspace*{1em} &\multicolumn{4}{c}{Proposed Methods}\\

\cline{3-5} \cline{7-10}

\multicolumn{1}{c}{Kernel Function}&{Algorithms}&\multicolumn{1}{c}{DDAG}&\multicolumn{1}{c}{ ADAG }&\multicolumn{1}{c}{Max Wins}&
&\multicolumn{1}{c}{SE}&\multicolumn{1}{c}{ RADAG }&\multicolumn{1}{c}{WE}&\multicolumn{1}{c}{VCF}\\
\hline
Polynomial
&DDAG          &&1-15-0	&2-14-0	&	&4-11-1	&5-9-2	&5-11-0	&6-10-0\\
&ADAG          &&		&2-13-1	&	&4-11-1	&5-9-2	&4-12-0	&5-11-0\\
&Max Wins      &&		&		&	&3-12-1	&5-9-2	&4-12-0	&4-12-0\\
&SE            &&		&		&	&		&2-14-0	&4-11-1	&4-11-1\\
&RADAG         &&		&		&	&		&		&2-12-2	&4-10-2\\
&WE           &&		&		&	&		&		&		&1-15-0\\
\hline
RBF
&DDAG          &&2-14-0	&2-14-0	&	&2-14-0	&3-13-0	&5-11-0	&5-11-0\\
&ADAG          &&		&2-14-0	&	&1-15-0	&3-13-0	&5-11-0	&5-11-0\\
&Max Wins      &&		&		&	&2-12-2	&1-14-1	&5-11-0	&5-11-0\\
&SE            &&		&		&	&		&2-13-1	&4-12-0	&3-13-0\\
&RADAG         &&		&		&	&		&		&3-13-0	&3-13-0\\
&WE           &&		&		&	&		&		&		&1-15-0\\
\hline
\end{tabular}}
\label{pair_comparison}
\end{table*}

Table~\ref{pair_comparison} summarizes paired comparisons of all algorithms including
the traditional techniques, and the proposed works based on both of the Polynomial kernel and
the RBF kernel. We show the win-draw-loss record (s) of the algorithm in the column
against the algorithm in the row. A win-draw-loss record reports how many datasets the method in the column 
is better than the method in the row (win), is equal (draw), or is worse (loss) at $95$\% confidence interval.
As summarized in the table, our proposed methods are better than all previous works i.e., the DDAG, the ADAG, and Max Wins.
WE and VCF give the highest accuracy among all of our methods.
The result also shows that VCF gives a little better results compared to WE.
However, as mentioned before in Section~\ref{VCF}, the accuracies of VCF are the ones without fine-tuning, and higher accuracies can be expected if fine-tuning is performed to find the optimal {\it $threshold\_value$} for VCF.

\vspace*{-0.5em}\subsection{Computational Time}\label{Time}
\begin{figure*}[htbp]
\centering
\vspace{-30pt}
\scalebox{0.4}{\includegraphics{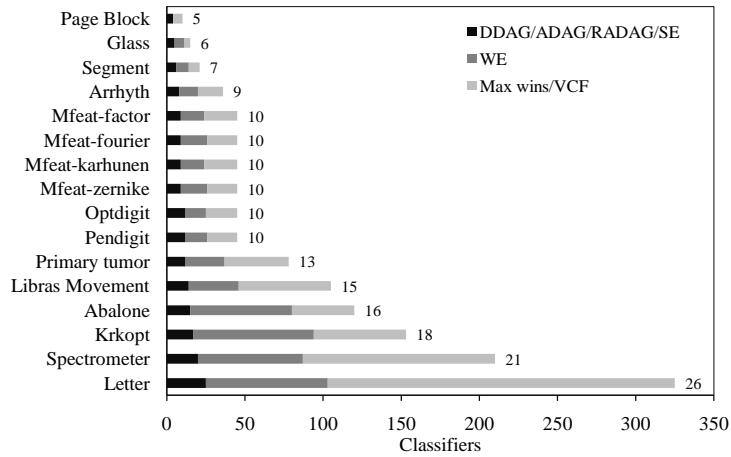}}
\vspace{-40pt}
\caption{A comparison of the computational time using the Polynomial kernel.}
\label{Time_poly}
\end{figure*}

\begin{figure*}[htbp]
\centering
\vspace{-30pt}
\scalebox{0.4}{\includegraphics{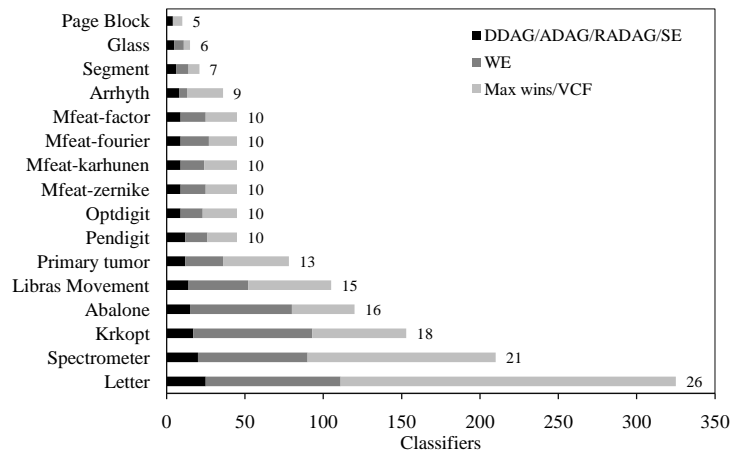}}
\vspace{-40pt}
\caption{A comparison of the computational time using the RBF kernel.}
\label{Time_rbf}
\end{figure*}

The computational times of all methods are shown in Fig.~\ref{Time_poly} and Fig.~\ref{Time_rbf}.
We can classify algorithms according to the time requirement into three groups, for an $N$-class problem:
1) $N-1$ times i.e., the DDAG, the ADAG, SE, and the RADAG,
2) average about half of time of $N(N-1)/2$ i.e., WE,
3) $N(N-1)/2$ i.e., Max Wins, and VCF.

The results show that algorithms in the first and the second groups
require comparatively low running time in all datasets, especially
when the number of classes is relatively large, while the larger the number of classes, the more running time the algorithms in the third group requires.
WE in the second group requires $N-1$ classifiers
in the best case and $N(N-1)/2$ classifiers in the worst case;
however, in our experimental results WE takes approximately half of time required by the algorithms in the third group. For the RADAG, though the number of classes affects the running time for reordering process, it takes a little time
even when there are many classes. The algorithms in the third group need $O(N^2)$ comparisons for a problem with $N$ classes.
VCF needs more time to choose the final class from the set of candidate classes
which can be obtained by re-using the previous results of binary classification.

The DDAG reduces the number of comparisons down to $O(N)$. SE spends a little time more than the DDAG
for sorting the classifiers in the training phase. By reducing the depth of the path,
the ADAG and SE require $O(N)$ comparisons of binary classifiers.
WE consumes more time than SE due to each round of classification can reduce only
one classifier while SE can eliminate all classifiers built from the discarded class.
The number of testing classifiers for WE is equal to that for Max Wins in the worst case; 
fortunately, the experimental results show that WE actually spends only half of Max Wins' times in the average case.
The RADAG needs a little time more than the ADAG for reordering the order of classes.
Note that, the minimum weight perfect matching algorithm, which is used in the
reordering algorithm, runs in time bounded by $O(N(M+N log N))$ \cite{Cook97}, where $N$
is the number of nodes (classes) in the graph and $M=N(N-1)/2$ is the number of edges (binary classifiers).
The RADAG will reorder the order of classes in every level, except for the last level.
The order of classes in the top level is reordered only once and we use the order to
evaluate every test example. Hence for classifying each test data, we need $log_2 N - 2$ times
of reordering, where each time the number of classes is reduced by half. Therefore, the running
time of the RADAG is bounded by $O(c_1N) +  O(c_2N^3 log_2 N)$, where $c_1$ is much larger than $c_2$.

\section{Conclusion}\label{conclusion}

Max Wins is a powerful combining technique with 
a need of $N(N-1)/2$ number of classifications for an $N$-class problem,
while the DDAG and the ADAG reduce the number of classifications to $N-1$. 
We study the characteristics of these previous methods that lead to wrong classification results.
We believe that the performances of them depend on the BCRTs.
In case of Max Wins, if there exists only one BCRT giving an incorrect answer, 
it may convey misclassification due to equal voting or another non-target class reaching the largest vote,
while in cases of the DDAG and the ADAG, if only one of BCRTs in the sequence of selected classifiers makes a mistake, 
the whole system will give the wrong output. We investigate the well-organized combination of the 
binary models including BCRTs in classification process to provide a more precise final result.

In this research, we propose four methods for overcoming the above weakness of the previous works. All our proposed methods are based on the same principle that if the information about genearalization ability is accurately measured, then it is able to be employed for enhancing the performance of the classification. In this paper, the generalization performance is estimated by $k$-fold cross-validation technique, and we show that it is more suitable than previously used measures in other frameworks, such as the margin size and the number of support vectors. Our proposed methods are the Reordering Adaptive Directed Acyclic Graph (RADAG), 
Strong Elimination of the classifiers (SE), Weak Elimination of the classifiers (WE), 
and Voting based Candidate  (VCF).
The RADAG is an enhanced version for the ADAG by using the minimum weight perfect matching for selecting 
the optimal pair of classes in each level with minimum generalization error. 
Compared to the ADAG, the RADAG is not only superior in terms of accuracy, but also maintains the same testing time ($N-1$). Next, We propose two improved algorithms for the DDAG, i.e. SE and WE. 
In SE, a sequence of binary classifiers selected by minimum generalization error 
is applied to eliminate the candidate classes until only one class remained and assigned 
as the final output class. SE provides better accuracy than the DDAG. 
The testing time of SE is the same as the traditional DDAG 
and the RADAG, with a number of applied classifiers equal to $N-1$. 
We also propose the other enhanced version for the DDAG, called WE.
This approach aims to efficiently use as many as possible of the classifiers with low generalization errors. This is different from the process of the DDAG and SE in which all binary classifiers related to a defeated class are ignored when the defeated class is removed from the candidate classes. In WE, however, a classifier will be ignored only if all of two related classed of that classifier are discarded from the candidate output classes, and this process enables WE to efficiently employ good classifiers.
WE gives significantly higher performance compared to the DDAG, and requires 
the number of classifications on average about half of the number of all possible binary classifiers.

Additionally, we propose VCF by applying the voting technique to carefully select 
the high competitive classes with high confidence. The remaining candidate classes 
are recursively eliminated by using WE. 
Although the number of classifications of VCF is equal to that of Max Wins, 
it shows the highest accuracy compared to all the other algorithms.
	
Finally, more experiments were conducted to compare our proposed algorithms 
and Max Wins in order to find the suitable scenario for using each of them.
The RADAG should be chosen when the number of classes is large and 
the classification time is the most concern. 
VCF shows the highest accuracy among our proposed algorithms, and it should be selected when the time constraints is not the main concern. In a general case, WE is the most suitable method 
because it is superior to Max Wins in terms of accuracy and time.
All of our techniques apply the generalization performance for organizing the use of the binary classifiers.
This measure can be optimally estimated by the mechanism of $k$-fold cross-validation that is independent of base learners. Consequently, all our proposed methods can be also applied to other base classifiers such as logistic regression, perceptron, linear discriminant analysis, etc.
The estimation of generalization errors using k-fold cross validation requires additional computation, and this can be thought of as a drawback of our methods. However, the estimation is done in the offline training phase, and thus it does not affect the performance in the classification phase.

\section{Acknowledgment}\label{Acknowledgment}
The authors would like to thank Dr.Peerapon Vateekul
for his valuable comments on an earlier version of this
paper. This research is partially supported by the Thailand Research Fund,
and the Graduate School, Chulalongkorn University.

\begin{wrapfigure}[7]{l}{2.4cm}
\vspace{-12pt}
  \scalebox{0.36}{\includegraphics{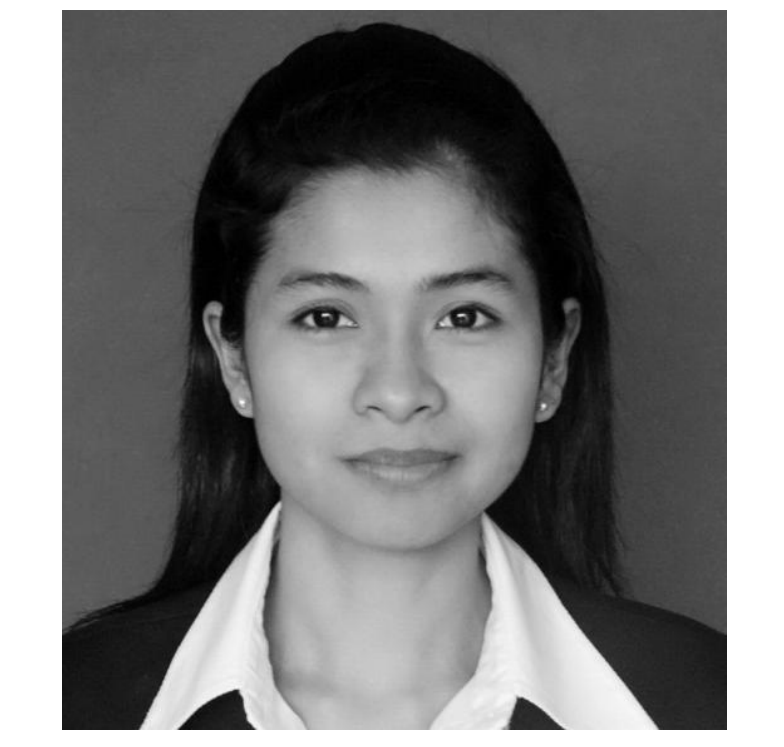}}         
\end{wrapfigure}

\hspace*{-2em}
{\footnotesize 
\textbf{Patoomsiri Songsiri} received the B.Sc. degree in Computer
Science (First class honor) from Prince of Songkla University, Thailand, in 2001
and the M.Sc. degree in Computer Science from Chulalongkorn University, Thailand, in 2006.
She is currently working toward the Ph.D. degree in 
Computer Engineering at Chulalongkorn University.
Her research interests include Pattern Recognition
and Machine Learning.\\ \\
}

\begin{wrapfigure}[7]{l}{2.4cm}
\vspace{-12pt}
  \scalebox{0.36}{\includegraphics{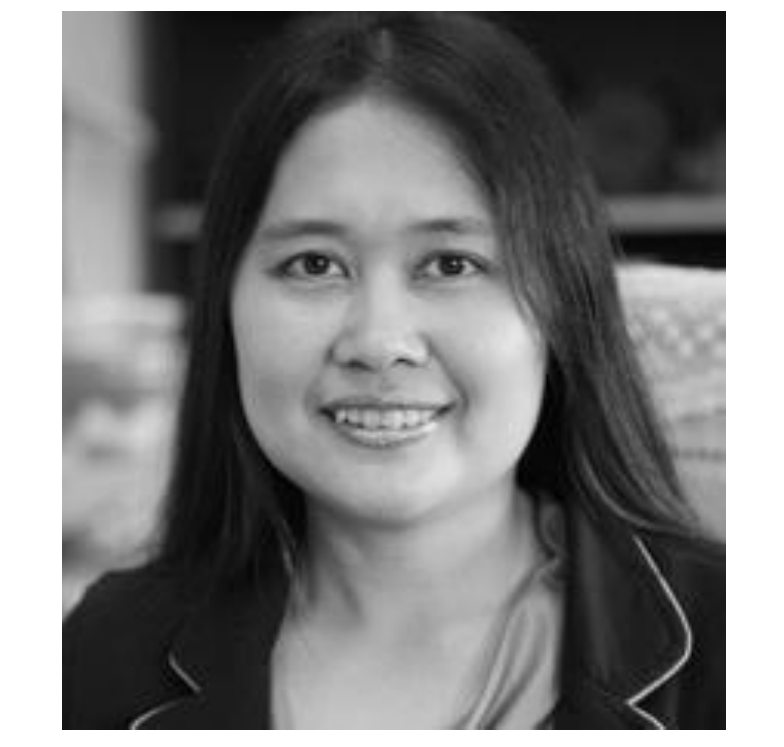}}          
\end{wrapfigure}
\hspace*{-2em}
{\footnotesize 
\textbf{Thimaporn Phetkaew} received her B.Sc. degree in Applied Mathematics 
and she also received her M.Sc. degree in Computer Science 
from Prince of Songkla University, Thailand in 1997 and 2000, respectively. 
In 2004, she received her Ph.D. degree in Computer Engineering from 
Chulalongkorn University, Thailand. Since 2004, as a lecturer, 
she has been with the School of Informatics, Walailak University. 
She is also a member of Informatics Innovation Research Unit 
at Walailak University. Her research interests include Data Mining, 
Machine Learning, and Software Testing.\\ \\
}

\begin{wrapfigure}[7]{l}{2.4cm}
\vspace{-12pt}   
 \scalebox{0.36}{\includegraphics{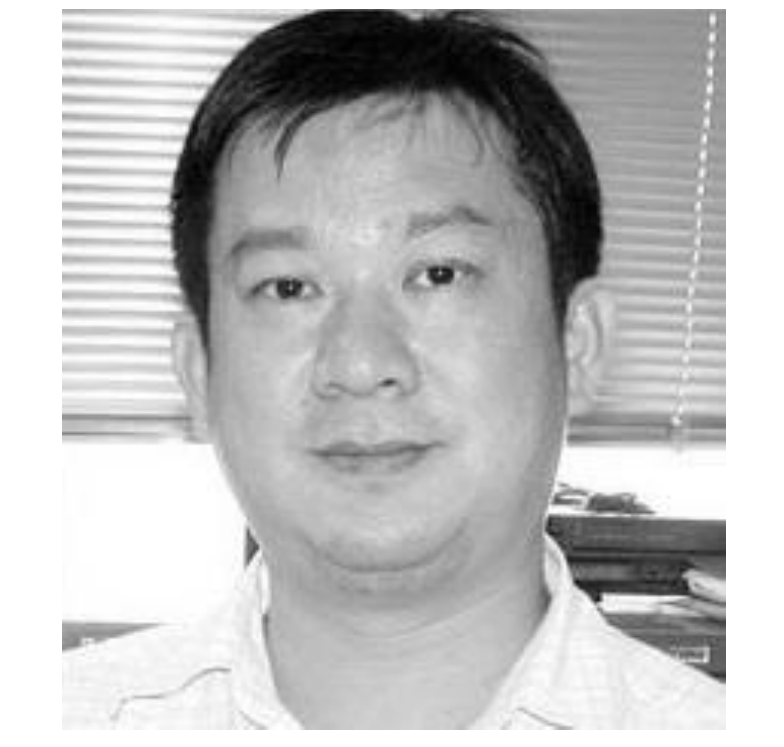}}      
\end{wrapfigure}
\hspace*{-2em}
{\footnotesize 
\textbf{Boonserm Kijsirikul} received the B.Eng. degree in Electronic and Electrical Engineering, 
the M.Sc. degree in Computer Science, and the Ph.D. in Computer Science 
from Tokyo Institute of Technology, Japan, in 1986, 1990, and 1993, respectively.
He is currently a Professor at the Department of Computer Engineering, Chulalongkorn University, Thailand. 
His current research interests include Machine Learning, Artificial Intelligence,
Natural Language Processing, and Speech Recognition.
}

\end{document}